\documentclass[10pt]{wlscirep}

\usepackage{caption}
\usepackage[utf8]{inputenc}
\usepackage[T1]{fontenc}
\usepackage{algorithm}
\usepackage{algorithmic}
\usepackage{multirow}
\usepackage{lineno}

\title{RPNT: Robust Pre-trained Neural Transformer - A Pathway for Generalized Motor Decoding}

\author[1]{Hao Fang}
\author[2]{Ryan A. Canfield}
\author[1]{Tomohiro Ouchi}
\author[2]{Beatrice Macagno}
\author[1,3,*]{Eli Shlizerman}
\author[1,2,4,*]{Amy L. Orsborn}

\affil[1]{Department of Electrical and Computer Engineering, University of Washington, Seattle, 98195, USA.}
\affil[2]{Department of Bioengineering, University of Washington, Seattle, 98195, USA}
\affil[3]{Department of Applied Mathematics, University of Washington, Seattle, 98195, USA}
\affil[4]{Washington National Primate Research Center, Seattle, 98195, USA}

\affil[*]{Authors to whom any correspondence should be addressed}
\affil[*]{shlizee@uw.edu and aorsborn@uw.edu }

\begin{abstract}
Brain motor decoding aims to interpret and translate neural activity into behaviors. Decoding models that are able to generalize across variations, such as recordings from different brain sites, experimental sessions, behavior types, and subjects, will be critical for real-world applications. Current decoding models only partially address these challenges. In this work, we develop a pretrained neural transformer model, RPNT - Robust Pretrained Neural Transformer, designed to achieve robust generalization through pretraining, which in turn enables effective finetuning for downstream motor decoding tasks. We achieved the proposed RPNT architecture by systematically investigating which transformer building blocks could be suitable for neural spike activity modeling, since components from models developed for other modalities, such as text and images, do not transfer directly to neural data.  The final resulting RPNT architecture incorporates three unique enabling components: 1) Multidimensional rotary positional embedding (MRoPE) to aggregate experimental metadata such as site coordinates, session ids and behavior types; 2) Context-based attention mechanism via convolution kernels operating on global attention to learn local temporal structures for handling non-stationarity of neural population activity; 3) Robust self-supervised learning (SSL) objective with stochastic causal masking strategies and contrastive representations. We pretrained two versions of RPNT on distinct datasets that present significant generalization challenges: a) Multi-session, multi-task, and multi-subject microelectrode benchmark; b) Multi-site recordings using high-density Neuropixel 1.0 probes from many cortical locations. Both datasets include recordings from the dorsal premotor cortex (PMd) and from the primary motor cortex (M1) regions of nonhuman primates (NHPs) performing reaching tasks. After pretraining, we evaluated RPNT generalization in terms of cross-session, cross-type, cross-subject, and cross-site downstream behavior decoding tasks. Our results show that RPNT consistently outperforms the decoding performance of existing decoding models on these tasks. Our ablation and hyperparameter sweep analyses demonstrate the necessity and robustness of the proposed novel components. Our work has implications for future robust brain decoder designs in brain-computer interfaces (BCIs). 

\end{abstract}
\begin{document}

\flushbottom
\maketitle

\thispagestyle{empty}

\section*{Introduction}
The transformer architecture~\cite{vaswani2017attention}, together with pretraining, has notably shifted data modeling paradigms, beginning in Natural Language Processing (NLP) and subsequently across many fields. Scalable transformer models such as BERT~\cite{devlin2019bert}, GPT~\cite{radford2019language}, and Vision Transformers~\cite{dosovitskiy2020image} demonstrated that a large scale pretraining on diverse data, followed by task specific finetuning, yields superior performance across related tasks compared with models trained for individual tasks. However, current neural modeling frameworks have not yet fully capitalized on transformer pretraining, in part because neural data exhibit session-dependent and nonstationary spatiotemporal structure. As an alternative, classical approaches leveraged dimensionality reduction to extract low-dimensional structure from high-dimensional session recordings. Methods such as Principal Component Analysis (PCA)~\cite{cunningham2014dimensionality} and Factor Analysis~\cite{santhanam2009factor} were applied to neural activity, which captured and identified dominant modes of variance. Such methods assume simple, static, and linear relationships. Thus, extensions have been developed, such as the recent probabilistic geometric PCA method for neural data application~\cite{hsieh2025probabilistic}  and Canonical Correlation Analysis (CCA) for modeling pairwise and cross-region interactions~\cite{semedo2019cortical}. While these methods extend classical dimension reduction approaches, they remain limited in capturing sequential neural properties. As an alternative, dynamical systems approaches such as Gaussian-Process Factor Analysis (GPFA)~\cite{yu2008gaussian} model the stochastic temporal evolution of neural latent states. These models have generally been applied to facilitate the interpretation of neural population data in offline analysis applications.

 In parallel, motor decoding models aim to reconstruct intended movement variables (e.g., kinematics) from neural population activity. Well-established approaches, such as Wiener~\cite{van2018signal} and Kalman filters~\cite{wu2002neural,orsborn2014closed}, provide computationally efficient real time decoding, but typically assume linear stochastic dynamics. Bayesian decoders~\cite{wu2006bayesian,shanechi2016robust} incorporate prior information about movement statistics, improving robustness but remaining constrained by parametric assumptions. Deep learning approaches have also been developed, progressing from MLPs~\cite{glaser2020machine} to RNN architectures~\cite{mante2013context,sani2024dissociative} that capture temporal dependencies in neural dynamics for effective decoding. More recent work explored Transformer-based decoders~\cite{ye2021representation,le2022stndt} that leverage self-attention to model long range dependencies across neural populations. However, these methods are typically trained from scratch on limited task specific data, resulting in poor generalization across sessions, subjects, and experimental conditions. To address this limitation, recent efforts explored large scale pretraining for neural data analysis using a variety of model architectures. Prior to transformer-based models, RNN-based approaches were pioneered and explored for multi-session pretraining, such as Multiplicative Recurrent Neural Network (MRNN)~\cite{sussillo2016making} and Latent Factor Analysis via Dynamical Systems (LFADS)~\cite{pandarinath2018inferring}. 
 
 More recent advances in transformer architecture models represent further progress towards pretrained neural models. In particular, Neural Data Transformer (NDT)~\cite{ye2021representation} introduced a transformer architecture for spike train modeling, while NDT2~\cite{ye2023neural} extended to multi context pretraining across sessions, subjects, and tasks using a spatiotemporal transformer. NDT3~\cite{ye2025generalist} scaled it to hundreds of datasets and showed emergent zero-shot/few-shot capabilities. Concurrently, supervised pretraining models such as Pre-training On manY neurOns (PoYo)~\cite{azabou2023unified, azabou2024multi} and PoSSM~\cite{ryoo2025generalizable} adopted a complementary approach that leverages learnable neural token representations to enable scalable pretraining across multiple subjects. Additional approaches, such as Population Transformer (PopT) used a standard transformer encoder architecture but with novel ensemble-wise and channel-wise discrimination tasks for training~\cite{chau2025population}. BrainBERT~\cite{wang2023brainbert} further adapts language model architectures by treating neural signals as text sequences. While effective, these transformer models lack explicit mechanisms to address non-stationarity and recording configuration variability in session based neural spike activity.

The aforementioned efforts show the promise of adapting pretraining strategies to neural datasets with larger samples of neural activity (i.e., neural population and neural dynamics models) and more diverse samples (distinct recording sessions and subjects). Yet,  robust generalization, via explainable mechanisms for variations in neural data, continues to present challenges, with several key aspects of variation in neural data remaining insufficiently addressed. First, model generalization remains limited when applied to unseen brain regions or different recording configurations~\cite{jude2022robust,karpowicz2024few,le2025spint,canfield2025spatiotemporal}. Second, neural signals exhibit drift over time~\cite{chestek2011long}. Decoding efforts will benefit from improved mechanisms to robustly handle neural nonstationarity. Third, existing approaches often emphasize training objectives that improve decoding performance through denoising neural signals, but may fail to learn representations that reflect the underlying causal dynamics that underly neural activity patterns~\cite{lu2025netformer}. Decoders that aim to learn causal brain-behavior relationships will likely be beneficial for data-driven discovery and real-time applications. 

\begin{figure}
    \centering
    \includegraphics[width=1\linewidth]{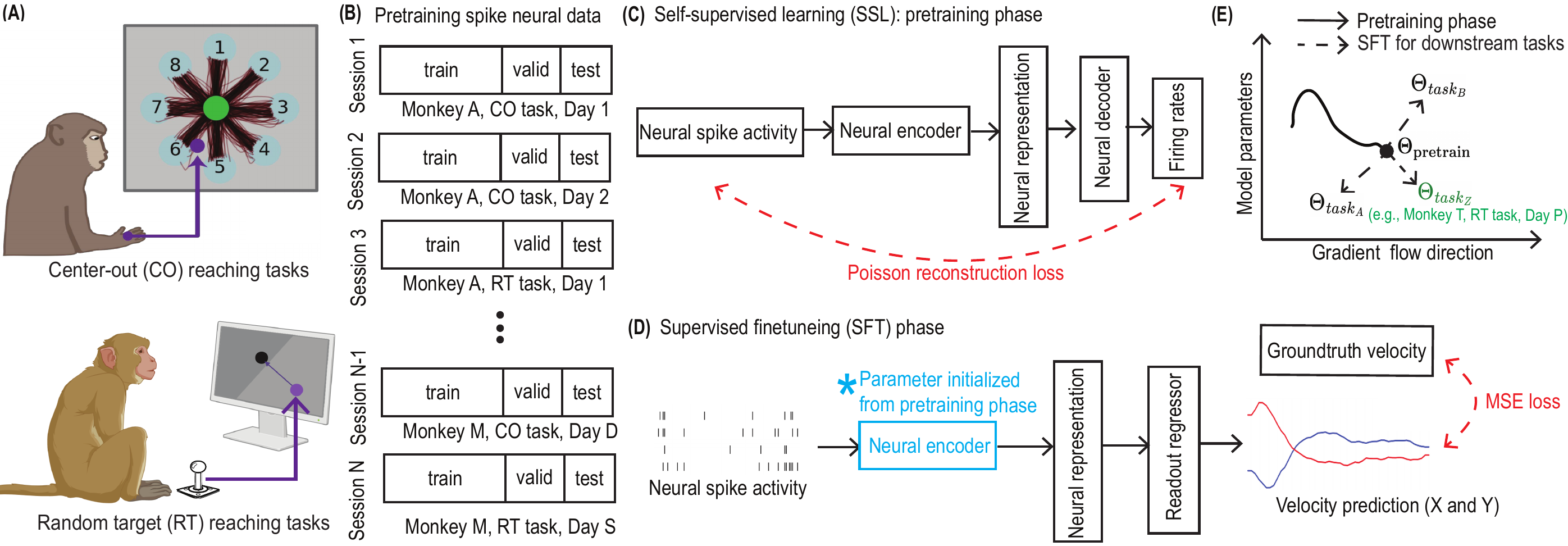}
    \caption{Overview of the pretraining and finetuning workflow for generalized motor decoding. (A) Experiments in which NHPs performed reaching tasks. (B) Preparation of training, validation, and test splits for the pre-training datasets. session. (C, D) Schematics for self-supervised learning (SSL) and downstream supervised finetuning (SFT), respectively. (E) Illustration of model parameter adaptation during the pretraining and finetuning phases.}
    \label{Figure 1: Background}
\end{figure}

We introduce RPNT - a neural transformer model that leverages robust pretraining enhancements to address the limitations of current approaches. RPNT contains novel components that emerged from our investigation of transformer architectures tailored for neural activity and its variations. The key novel components of RPNT are:  1) Multidimensional rotary positional embedding (MRoPE) that aggregates experimental metadata such as site coordinates, session ids, and behavior types; 2) A context-based attention mechanism that applies convolution kernels to global attention to learn local temporal structure and handle the non-stationarity of neural activity; 3) A robust self supervised learning (SSL) objective with stochastic causal masking strategies and contrastive representation learning. For experimental validation, we followed and implemented a standard pretraining and finetuning workflow (see Figure~\ref{Figure 1: Background}). Using this workflow, we pretrained two versions of RPNT on two datasets containing different modalities of recorded neural activity: a) A multi-session, multi-task, and multi-subject microelectrode benchmark data~\cite{perich2018neural}; b) A multi-site recordings using high-density Neuropixel 1.0 probes sampling neural activity across many cortical locations~\cite{canfield2025spatiotemporal}. Both datasets include recordings from PMd and M1 regions of nonhuman primates (NHPs) as they performed center-out or random target reaching tasks. After pretraining, we evaluated the generalization of RPNT on cross-session, cross-type, cross-subject, and cross-site behavioral decoding tasks. RPNT consistently matched or exceeded the decoding performance of existing decoding models across all tasks. Ablation and hyperparameter analyses demonstrated the necessity and robustness of the proposed novel components. Furthermore, we observed that the learned spatial attention maps provide data-driven insights into the structure of neural encoding of movement variables. Overall, our investigations and the novel components of RPNT support future large-scale neural transformer pretraining for decoding applications such as brain computer interfaces (BCIs)~\cite{robinson2025application} and data-driven discovery for neuroscience~\cite{canfield2025spatiotemporal}. To this end, our main contributions are:
\begin{itemize}  
    \item \textbf{Model architecture}: We propose a Multidimensional Rotary Positional Embedding (MRoPE) to aggregate experimental metadata for decoding generalization and introduce a context-based attention mechanism to handle the non-stationarity. 
    \item \textbf{Pretraining strategy}: We pretrain RPNT using a robust self-supervised learning objective via stochastic causal masking strategies and contrastive representations.
    \item \textbf{Generalized Decoding}: We demonstrate superior decoding performance for 1) cross-session, cross-behavior types, and cross-subject scenarios in the microelecotrode recording benchmark; 2) cross-site scenarios in a new Neuropixels dataset. 
\end{itemize}

\section*{Results}
\subsection*{Datasets and data collection}
\textbf{Long-term recordings of motor and premotor cortical spiking activity during reaching in monkeys (LTRCH) benchmark dataset}
We evaluated the downstream generalization capabilities of the RPNT in the presence of three forms of neural variability captured by an existing benchmark dataset: "\textit{Long-term recordings of motor and premotor cortical spiking activity during reaching in monkeys}" ~\cite{dandi:DANDI:000688_0.250122.1735} (LTRCH). The dataset allows evaluation of cross-session, cross-task, and cross-subject generalization, as described in prior work~\cite{perich2018neural}. The LTRCH dataset contains electrophysiology and behavioral data from $4$ rhesus macaques (subjects C, J, M, and T) performing a center-out reaching task (denoted as CO) or a continuous random target acquisition task (denoted as RT). Neural activities were recorded from chronically-implanted electrode arrays in the primary motor cortex (M1) or dorsal premotor cortex (PMd). LTRCH contains 111 sessions, spanning 43 recording hours. Both raw data and preprocessed data can be accessed from the Python package \textit{brainsets}~\cite{azabou2023unified}. 

\begin{figure}[t]
    \centering
    \includegraphics[width=0.7\linewidth]{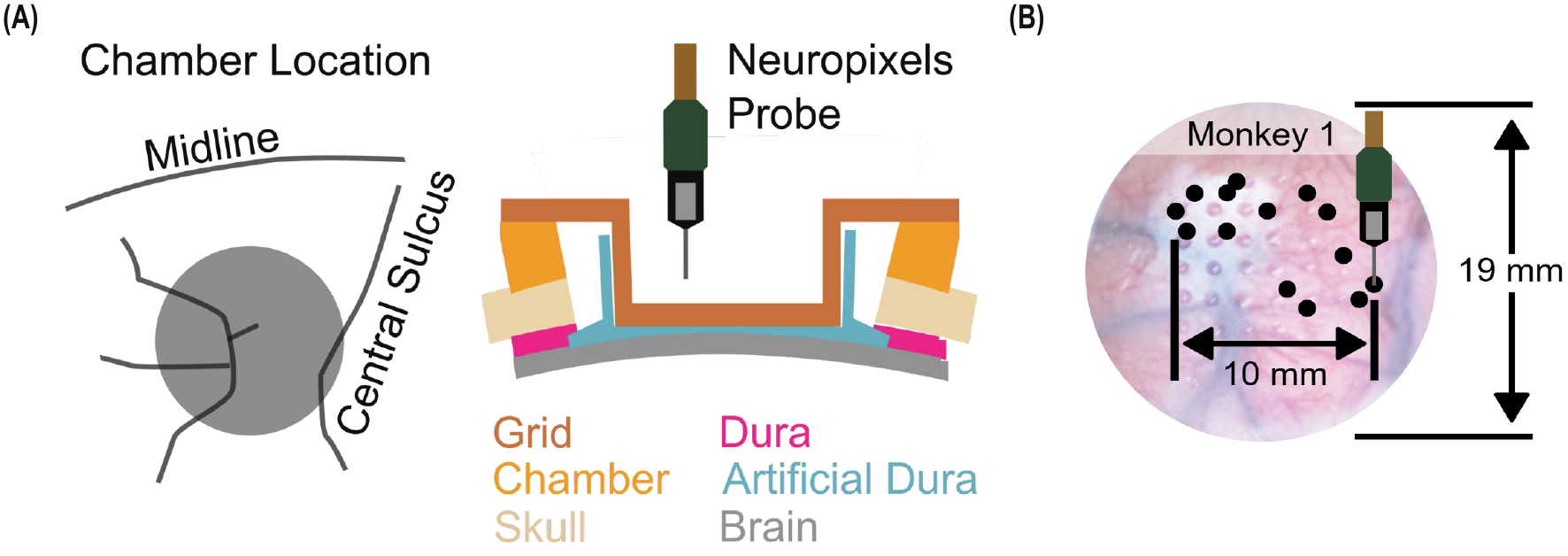}
    \caption{Neuropixel data collection setup. (A) Chamber location and neuropixel probe. (B) Neuropixels probe insertion locations (black dots) on the cortical surface. Each black dot stands for a different recording site.}
    \label{Figure 2: Neuropixel setup}
\end{figure}

\textbf{Neuropixel dataset (NPCS)}
We evaluated whether RPNT can also generalize across measurement location within the brain (cross-site) using a newly-generated dataset: "\textit{NeuroPixels across Cortical Sites}" (NPCS) ~\cite{canfield2025spatiotemporal}. The dataset allows evaluation of cross-site generalization within a single behavioral task and subject. The NPCS dataset includes electrophysiology and behavioral data from one male rhesus macaque (subject B) performing a center-out reaching task (denoted as CO). In each experimental session, a single neuropixel probe was inserted into the cortex via an implanted recording chamber to measure neural activity from 384 electrodes simultaneously. The probe insertion location was varied across sessions to span over 1 cm of M1 and PMd (see Neuropixel probe setup in Figure~\ref{Figure 2: Neuropixel setup} and Appendix Table~\ref{table: site_breakdown}). The dataset is comprised of 17 recordings at different recording sites collected during experiment sessions spanning approximately one year (see Appendix~\ref{Appendix: preprocessing}). The NPCS dataset was collected at the University of Washington and the Washington National Primate Research Center. All procedures were conducted in compliance with the NIH Guide for the Care and Use of Laboratory Animals and were approved by the Institutional Animal Care and Use Committee at the University of Washington. Details of the Neuropixel recording setup and data collections are further described in our prior work~\cite{canfield2025spatiotemporal}.

\subsection*{RPNT pretraining and evaluation}
We conducted experiments to evaluate RPNT, first focusing on self-supervised pretraining and then on downstream decoding generalization through supervised finetuning (SFT) with baseline comparisons on test recordings (see Figure~\ref{Figure 1: Background}E for model adaptation). In LTRCH~\cite{dandi:DANDI:000688_0.250122.1735}, we followed the setup in POSSM~\cite{ryoo2025generalizable}. We used designated training sessions from subjects C, J, and M for pretraining, and the remaining sessions for downstream evaluation. Our decoding evaluation included three cases: 1) New center out sessions from Monkey C (denoted as C-CO) for cross session evaluation; 2) New center-out sessions from Monkey T (denoted as T-CO) for cross subject evaluation; 3) New random target sessions from Monkey T (denoted as T-RT) for cross subject and cross tasks evaluation. Since more than $80\%$ of sessions in the LTRCH benchmark are center-out tasks, T-RT is the most challenging evaluation, since it requires adaptation to both subject and task distribution shifts. We followed the released code and customized it to build the train/validation/test datasets according to the Python package \textit{torch brain} ~\cite{azabou2023unified}. Details of the processing are provided in the Appendix~\ref{Supplementary: Data Pre-processing Pipeline}. For NPCS, we used 16 sites (S1-S16) for pretraining (0.8/0.1/0.1 for train/validation/test split), and the remaining site (S17) was used for Few-Shot (FS) SFT of Cross Site (CS) evaluation (0.2/0.3/0.5 train/validation/test split). As the CO task is highly structured, we evaluated decoding performance during the reaching period on successful trials following prior work~\cite{ryoo2025generalizable}. For RT tasks, we evaluated the performance using the whole period for all trials. We used $R^{2}$ to quantify decoding performance between predicted and ground truth velocity. Baseline comparison models are listed in the Appendix~\ref{Supplementary: Compare with Baseline Models}.

\subsection*{Cross-session, cross-subject, and cross-task evaluation results on LTRCH}
\begin{table}[tb]
\small
\centering
\begin{tabular}{l|l|c|c|c}
\hline
 & Method & Cross-Session (C-CO) & Cross-Subject (T-CO) & Cross-Task (T-RT) \\
\hline
\multirow{9}{*}{\rotatebox{90}{From scratch}} 
& Wiener filter~\cite{van2018signal} & 0.8712 $\pm$ 0.0137 & 0.6597 $\pm$ 0.0392 & 0.5942 $\pm$ 0.0564 \\
& MLP $^{\dagger,}$~\cite{glaser2020machine} & 0.9210 $\pm$ 0.0010 & 0.7976 $\pm$ 0.0220 & 0.7007 $\pm$ 0.0774 \\
& S4D $^{\dagger,}$~\cite{gu2021efficiently} & 0.9381 $\pm$ 0.0083 & 0.8526$\pm$ 0.0243 & 0.7145$\pm$ 0.0671 \\
& Mamba $^{\dagger,}$~\cite{gu2023mamba} & 0.9287 $\pm$ 0.0034 & 0.7692$\pm$ 0.0235 & 0.6694$\pm$ 0.1220 \\
& GRU $^{\dagger,}$~\cite{cho2014learning} & 0.9376$\pm$ 0.0036 & 0.8453$\pm$ 0.0200 & 0.7279$\pm$ 0.0679 \\
& POYO-SS $^{\dagger,}$~\cite{azabou2023unified} & 0.9427$\pm$ 0.0019 & 0.8705$\pm$ 0.0193 & 0.7156$\pm$ 0.0966 \\
& POSSM-S4D-SS $^{\dagger,}$~\cite{ryoo2025generalizable} & 0.9515$\pm$ 0.0021 & 0.8838$\pm$0.0171 & 0.7505$\pm$ 0.0735 \\
& POSSM-Mamba-SS $^{\dagger,}$~\cite{ryoo2025generalizable} & 0.9550$\pm$ 0.0003 & 0.8747$\pm$0.0173 & 0.7418$\pm$ 0.0790 \\
& POSSM-GRU-SS $^{\dagger,}$~\cite{ryoo2025generalizable} & 0.9549$\pm$ 0.0012 & 0.8863$\pm$0.0222 & 0.7687$\pm$ 0.0669 \\
& \textbf{RPNT} & \textbf{0.9647}$\pm$ 0.0026 & \textbf{0.9103}$\pm$0.0182 & \textbf{0.8356}$\pm$ 0.0914\\
\hline
\hline
\multirow{7}{*}{\rotatebox{90}{Pretrained}}
& NDT-2 (FT) $^{\dagger, *,}$~\cite{ye2023neural} & 0.8507$\pm$0.0110 & 0.6549$\pm$ 0.0290 & 0.5903$\pm$0.1430 \\
& POYO-1 (FT) $^{\dagger, *,}$~\cite{azabou2023unified}  & 0.9611$\pm$0.0035 & 0.8859$\pm$0.0275 & 0.7591$\pm$0.0770 \\
& o-POSSM-S4D (FT) $^{\dagger, *,}$~\cite{ryoo2025generalizable} & 0.9618 $\pm$ 0.0007 & 0.9069$\pm$ 0.0120 & 0.7584$\pm$ 0.0637 \\
& o-POSSM-Mamba (FT) $^{\dagger, *,}$~\cite{ryoo2025generalizable}  & 0.9574$\pm$ 0.0016 & 0.9011$\pm$ 0.0148 & 0.7621$\pm$ 0.0765 \\
& o-POSSM-GRU (FT) $^{\dagger, *,}$~\cite{ryoo2025generalizable} & 0.9587$\pm$ 0.0052& 0.9021$\pm$ 0.0241 & 0.7717$\pm$ 0.0595 \\
& \textbf{RPNT} (FS-SFT) & \textbf{0.9801}$\pm$ 0.0060 & \textbf{0.9431}$\pm$0.0103 & \textbf{0.8515}$\pm$0.1071 \\
\hline
& \textbf{RPNT} (Full-SFT) & \textbf{0.9894}$\pm$ 0.0037 & \textbf{0.9626}$\pm$0.0059 & \textbf{0.8778}$\pm$0.1005 \\
\hline
\end{tabular}

\textcolor{black}{$^{\dagger}$ These results were obtained and reported in recent POSSM work~\cite{ryoo2025generalizable} and cited in this table for comparison. $^{*}$ The pretrained section includes the Full Model Finetuning (FT) results~\cite{ryoo2025generalizable}, which are directly related to the RPNT experimental setup.}
\caption{Velocity decoding performance ($R^2$) comparison across three generalization scenarios (C-CO, T-CO, and T-RT) on the LTRCH benchmark. RPNT (FS-SFT) uses few shot finetuning, whereas RPNT (Full-SFT) uses all available downstream sessions. Mean and standard deviation are computed across sessions. Bold numbers indicate the best performance in each scenario. References for baseline models are provided in Appendix~\ref{Supplementary: Compare with Baseline Models} and comments $^{\dagger}$ and $^{*}$.}
\label{table: Cross-session, cross-subject, and cross-task evaluation results in public dataset}
\end{table}

We compared velocity decoding performance ($R^2$) across three generalization scenarios on the LTRCH benchmark (Table~\ref{table: Cross-session, cross-subject, and cross-task evaluation results in public dataset}).  RPNT was evaluated under two training settings: (1) models trained from scratch on single downstream sessions and (2) pretrained models finetuned with different amounts of downstream data.

\textbf{Training from scratch.}  We first trained models from scratch using a single downstream session. We compared previously published models with RPNT following the training and evaluation approach used in recent work (POSSM) ~\cite{ryoo2025generalizable} (Table~\ref{table: Cross-session, cross-subject, and cross-task evaluation results in public dataset}, top). RPNT achieved the best performance across all three generalization challenges (cross-session, cross-subject, and cross-task). RPNT obtained $R^2$ scores (mean $\pm$ std) of $0.9647 \pm 0.0026$, $0.9103 \pm 0.0182$, and $0.8356 \pm 0.0914$ in C-CO, T-CO, and T-RT evaluations, respectively. These results demonstrated improved performance over existing decoding models, with a notable gain in the challenging T-RT scenario (approximately $7\%$ over the second-best baseline POSSM-GRU).

\textbf{Pretrained models.} We next investigated the benefits of model pretraining paired with downstream supervised finetuning (SFT) and compared performance in this regime across models. As a self-supervised pretraining model, we evaluated RPNT in two regimes: (a) finetuning on the full downstream training dataset (Full-SFT), and (b) few-shot finetuning on a single downstream session (FS-SFT). While we evaluated RPNT in both regimes, we compared its performance with that of other previously published models evaluated in the \textbf{FS-SFT} setting using existing evaluations (provided in POSSM~\cite{ryoo2025generalizable}; Table~\ref{table: Cross-session, cross-subject, and cross-task evaluation results in public dataset}, bottom). Our results showed that both RPNT (Full-SFT) and RPNT (FS-SFT) outperformed existing baselines, obtaining average $R^2$ scores of $0.9801$, $0.9431$, and $0.8515$. This was despite the fact that both RPNT regimes did not have any access to behavioral labels at any stage (both pretraining and testing), while existing baselines (e.g., POYO~\cite{azabou2023unified} and POSSM~\cite{ryoo2025generalizable}) in the FS-SFT setting had access to behavior labels from all sessions during pretraining. This shows that RPNT performs well in the computationally efficient and more constrained training settings imposed by the FS-SFT regime. We therefore focused solely on this setting in subsequent experiments (Section~\ref{Cross-site evaluation result on the Neuropixel dataset} and Section~\ref{Ablation Studies}). RPNT Performance was higher in the Full-SFT regime, compared to FS-SFT, producing $R^2$ scores (mean $\pm$ std) 0.9894 $\pm 0.0037$ (C-CO), $0.9626 \pm 0.0059$ (T-CO), and $0.8778 \pm 0.1005$ (T-RT) across tasks, showing that RPNT was also able to leverage access to multiple sessions to improve generalization. Overall, we found that RPNT consistently surpassed existing models, demonstrating improved efficiency and robustness in decoding generalization.

\subsection*{Cross-site evaluation result on the Neuropixel dataset}
\label{Cross-site evaluation result on the Neuropixel dataset}

\begin{figure}[ht]
\centering
\begin{minipage}{0.48\textwidth}
    \centering
    \small
    \begin{tabular}{l|c}
    \hline
    Method & Cross-Site (B-CS)  \\
    \hline
    Wiener filter~\cite{van2018signal} & 0.3462 $\pm$ 0.0710 \\
    MLP~\cite{glaser2020machine} &  0.4074 $\pm$ 0.0592  \\
    RNN (LFADS)~\cite{pandarinath2018inferring} & 0.5015 $\pm$ 0.1085  \\
    Transformer (NDT)~\cite{ye2021representation} & 0.5272 $\pm$ 0.0720  \\ 
    \textcolor{black}{Transformer (PoYo)}~\cite{azabou2023unified} & \textcolor{black}{0.5944} $\pm$ \textcolor{black}{0.0901}  \\
    \textbf{RPNT} (from scratch)& \textbf{0.6358} $\pm$ 0.0311 \\
    \textbf{RPNT} (pretrained) & \textbf{0.6612} $\pm$ 0.0328 \\
    \hline
    \end{tabular}
    \captionof{table}{Decoding performance comparison across sites (B-CS) on the neuropixel dataset. We train the baseline models from scratch on the NPCS dataset. The mean and standard deviation were calculated across all successful trials.}
    \label{table: cross-site in neuropixel dataset}
\end{minipage}
\hfill
\begin{minipage}{0.4\textwidth}
    \centering
    \includegraphics[width=\textwidth]{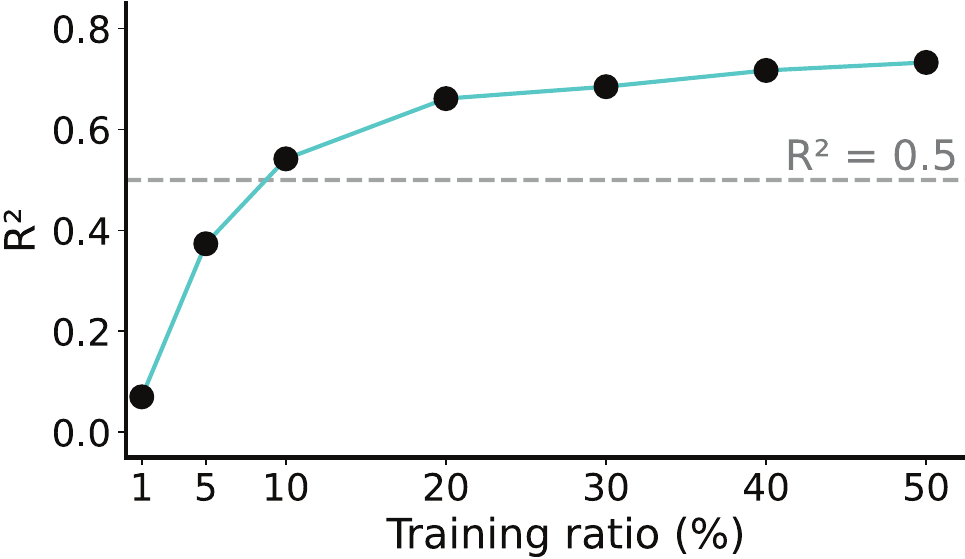}
    \vspace{-0.6cm}
    \caption{Sweep analysis for training splits for pretrained RPNT model.}
    \label{Figure 3: Result sweep train ratio}
\end{minipage}
\end{figure}

We further investigated the generalization of RPNT under Cross Site variations using the NPCS dataset. This dataset differs from the LTRCH benchmark in recording modality (Neuropixels vs. microelectrodes) and is particularly suitable for cross site analysis, as recordings do not share neurons across sites (see Figure~\ref{Figure 2: Neuropixel setup} for recording site topology). We compared decoding performance on RPNT with previously published model architectures with available implementation code (Table~\ref{table: cross-site in neuropixel dataset}). RPNT, in the FS-SFT regime, outperformed all existing benchmark model architectures on the downstream single site (S17) velocity decoding task. Furthermore, pretraining consistently improved performance (scratch: 0.6358 vs. pretrained: 0.6612).

We also swept the amount of data used for training (training split) from $1\%$ to $50\%$ ($20\%$ is set as default) while fixing the amount of data used for testing (test split) to $50\%$. We observed that even with $10\%$ training data (half of the default $20\%$ FS-SFT  train split), the pretrained RPNT model still maintained reasonable performance ( Figure~\ref{Figure 3: Result sweep train ratio}, slightly above 0.5 $R^{2}$), demonstrating its potential uses for applications with limited downstream data. RPNT outperformed other benchmark models tested across the majority of training splits assessed, with the exception of the extreme low-data regime of $1\%$ (Table~\ref{table: Sweep analysis of the train split ratio in the neuropixel dataset}). 

\begin{table}[h]
\centering
\small
\begin{tabular}{l|c|c|c|c|c|c|c}
\hline
Method & 1\% & 5\% & 10\% & 20\% & 30\% & 40\% & 50\%  \\
\hline 
Wiener filter & 0.0009 & 0.0344 & 0.0892 & 0.3462 & 0.3567 & 0.3597 & 0.3634 \\
MLP & 0.1181 & 0.3070 & 0.3704 & 0.4074 & 0.4215 & 0.4279 & 0.4314 \\
RNN & -0.0049 & -0.0215 & 0.1083 & 0.5015 & 0.5534 & 0.6229 & 0.6295 \\
NDT & -0.0038 & 0.0836 & 0.2401 & 0.5272 & 0.5752 & 0.6082 & 0.6175 \\
PoYo & -0.0136 & 0.2285 & 0.5378 & 0.5944 & 0.6578 & 0.6938 & 0.7285 \\
\hline 
RPNT (from scratch) & -0.4101 & 0.1553 & 0.4884 & 0.6358 & 0.6883 & 0.6969 & 0.7265 \\
RPNT (pretrained) & 0.0701 & \textbf{0.3735} &  \textbf{0.5418} &  \textbf{0.6612} &  \textbf{0.6927} &  \textbf{0.7170} &  \textbf{0.7329} \\
\hline
\end{tabular}
\caption{Sweep analysis of the train split ratio on the neuropixel dataset to show the robustness of the RNPT with limited downstream train data. The test split data consistently uses the last 50$\%$ trials. We report the mean $R^{2}$ score for clear comparison across different baselines.}
\label{table: Sweep analysis of the train split ratio in the neuropixel dataset}
\end{table}

\subsection*{Transfer datasets performance of our pretrained RPNT model}
\begin{table}
\centering
\begin{tabular}{l|c|c|c}
\hline
Method & C-CO & T-CO & T-RT \\
\hline 
Transfer from NPCS to LRTCH& 0.8603 $\pm$ 0.0009 & 0.7585 $\pm$ 0.0344  & 0.6395 $\pm$ 0.0892 \\
No transfer, only on LRTCH & 0.9894 $\pm$ 0.0037 & 0.9626 $\pm$ 0.0059  & 0.9626 $\pm$ 0.0059\\
\hline
\end{tabular}
\caption{Evaluation results of the NPCS pretrained RNPT model on the LRTCH dataset}
\label{table: Model Transfer from neuropixel to public}

\hfill 

\centering
\begin{tabular}{l|c}
\hline
Method & B-CS \\
\hline 
Transfer from LRTCH to NPCS& 0.6537 $\pm$ 0.0413\\
No transfer, only at NPCS & 0.6612 $\pm$ 0.0328 \\
\hline
\end{tabular}
\caption{Evaluation results of the LRTCH pretrained RNPT model on the NPCS dataset}
\label{table: Model transfer from public to neuropixel}
\end{table}

Both datasets consist of nonhuman primates performing similar motor tasks with recordings from similar brain regions, which gave us the opportunity to perform two experiments to evaluate transfer generalization of pretrained RPNT across datasets. First, we applied the NPCS pretrained RPNT model to the LTRCH dataset. We observed that RPNT had limited generalization performance (Table~\ref{table: Model Transfer from neuropixel to public}). In contrast, when we reversed the transfer direction by applying RPNT pretrained on the LTRCH dataset to the NPCS dataset, we observed that RPNT maintained strong generalization performance (transfer: 0.6537 vs. original: 0.6612) (Table~\ref{table: Model transfer from public to neuropixel}). These asymmetric results suggested that pretraining on the LTRCH dataset enabled more robust neural representations that could transfer effectively to the NPCS Neuropixel dataset. One possible explanation was the difference in data scale, where the LTRCH dataset contained more than 100 sessions, in contrast with 16 sessions on the NPCS dataset. These results further suggested the scalability benefits of leveraging larger datasets during the pretraining phase of RPNT. 

\subsection*{Ablation studies}
\label{Ablation Studies}

The RPNT model architecture contains multiple novel components. We next evaluated how each component contributed to RPNT's decoding performance using ablation studies. We conducted ablations of the positional encoding, attention mechanism, and masking strategy evaluated on the T-RT and B-CS tasks in the FS-SFT regime (Tables~\ref{table: ablation pe}, Tables~\ref{table: ablation attention}, and Tables~\ref{table: ablation masking strategy}). We performed additional sweeping analyses (e.g., transformer layers and attention heads), summarized in the Appendix (~\ref{Supplementary: Sweeping analysis for RPNT model}).

\textbf{Positional encoding.} We compared RPNT's positional encoding approach, MRoPE, with three alternative positional encoding approaches: sinusoidal PE, standard RoPE, and multi-dimensional learnable PE (Table~\ref{table: ablation pe}). MRoPE achieved the highest decoding performance ($R^2$ score). An improvement of approximately $3\%$ over standard RoPE supported the extension of RoPE to multiple dimensions, enabling generalization to session configurations while preserving relative position information.

\textbf{Context-based Attention mechanism.} We compared RPNT's context-based attention design with standard self-attention mechanisms (Table~\ref{table: ablation attention}). This comparison demonstrated the importance of the context-based attention design, as we observe a substantial performance drop of approximately $5\%$ when shifting to standard self-attention. This gap highlights the effectiveness of the adaptive kernel mechanism in handling the nonstationary nature of neural recordings. We further include a sweep study over different kernel sizes, in which we observed that context-based attention also consistently performed better than the standard attention. (see Appendix Table~\ref{table: Sweeping study on kernel size}).

\textbf{Random uniform masking strategy.} We compared RPNT's random uniform masking strategy to multiple fixed masking ratio strategies, including symmetric and asymmetric across neuron and temporal dimensions. All fixed masking ratio strategies consistently underperformed the random uniform masking strategy (Table~\ref{table: ablation masking strategy}). This result suggested that exposing the model to a range of masking ratios leads to more robust neural representations and improved downstream performance.

Together, these ablation studies confirmed that each proposed component contributed to the strong performance of RPNT in generalized motor decoding.

\begin{table}
\centering
\begin{tabular}{l|c|c}
\hline
Method & Cross-Task (T-RT) & Cross-Site (B-CS) \\
\hline 
Sinusoidal PE & 0.8260 $\pm$ 0.0894 & 0.6242 $\pm$ 0.0267 \\
RoPE & 0.8226 $\pm$ 0.0941 & 0.6484 $\pm$ 0.0074 \\
Learnable PE & 0.8305 $\pm$ 0.0917 & 0.6273 $\pm$ 0.0398 \\ 
\textbf{MRoPE} & \textbf{0.8515} $\pm$ 0.1071 & \textbf{0.6612} $\pm$ 0.0328 \\
\hline
\end{tabular}
\caption{Ablation study of positional encoding methods.}
\label{table: ablation pe}

\hfill 

\centering
\begin{tabular}{l|c|c}
\hline
Method & Cross-Task (T-RT) & Cross-Site (B-CS) \\
\hline 
Standard attention& 0.8024 $\pm$ 0.0896 & 0.5024 $\pm$ 0.0388 \\
\textbf{Context-based attention} & \textbf{0.8515} $\pm$ 0.1071 & \textbf{0.6612} $\pm$ 0.0328\\
\hline
\end{tabular}
\caption{Ablation study of attention mechanisms.}
\label{table: ablation attention}

\vspace{0.2cm}

\centering
\begin{tabular}{l|c|c|c|c}
\hline
& \multicolumn{2}{c|}{Masking strategy} & \multirow{2}{*}{T-RT} & \multirow{2}{*}{B-CS} \\
\cline{2-3}
& Neuron ratio & Temporal ratio  &  \\
\hline 
\multirow{5}{*}{Fixed} 
& 0.25 & 0.25 & 0.8349 $\pm$ 0.0948 & 0.6549 $\pm$ 0.0254 \\
& 0.50 & 0.50 & 0.8437 $\pm$ 0.0968 & 0.6557 $\pm$ 0.0316\\
& 0.75 & 0.75 & 0.8392 $\pm$ 0.0975 & 0.6593 $\pm$ 0.0251 \\
& 0.25 & 0.75 & 0.8414 $\pm$ 0.0990 & 0.6540 $\pm$ 0.0279 \\
& 0.75 & 0.25 & 0.8398 $\pm$ 0.0922 & 0.6594 $\pm$ 0.0301 \\ 
\hline
\multirow{1}{*}{\textbf{Random}} 
& \textbf{$\mathcal{U}(0, 1)$} & \textbf{$\mathcal{U}(0, 1)$} & \textbf{0.8515} $\pm$ 0.1071  & \textbf{0.6612} $\pm$ 0.0328 \\
\hline
\end{tabular}
\caption{Ablation study of masking strategies during pretraining.}
\label{table: ablation masking strategy}
\end{table}

\section*{Discussion and Conclusions} 

We present RPNT, a robust pretrained neural transformer designed for generalized motor decoding across sessions, subjects, behavior types, and recording sites. RPNT combines three novel components --- multidimensional rotary positional embedding (MRoPE), context-based attention mechanism, and uniform random masking strategy -- with a self-supervised learning (SSL) framework that eliminates the need for behavioral labels during pretraining. Unlike existing pretrained neural decoders such as POYO and its subsequent work ~\cite{azabou2023unified,azabou2024multi,ryoo2025generalizable}, which rely on supervised objectives that require behavior labels during pretraining, RPNT learns robust neural representations purely from spike activity and adapts to downstream tasks via few shot supervised finetuning (FS-SFT). We validated RPNT on two datasets: a multi-session, multi-subject, multi-task microelectrode benchmark~\cite{perich2018neural} (LTRCH) and a new multi-site Neuropixel 1.0 dataset~\cite{canfield2025spatiotemporal} comprising 17 recording sites spanning large portions of PMd and M1 over approximately one year (NPCS). The NPCS dataset evaluated in this work enabled the first cross site generalization benchmark for pretrained neural decoders. Across all four evaluation scenarios, RPNT consistently surpassed existing state-of-the-art models, notably outperforming supervised pretrained baselines that had access to behavior labels throughout pretraining. These results, together with comprehensive ablation studies, establish RPNT as a promising pathway toward neural foundation models for brain decoding.

The algorithmic innovations in RPNT directly address key limitations in existing neural transformer architectures. First, standard positional encodings, such as sinusoidal~\cite{vaswani2017attention} and learnable~\cite{devlin2019bert}, cannot represent the multi-dimensional experimental metadata inherent to neural recordings, such as electrode coordinates, session identities, and behavior types. MRoPE extends rotary positional embedding~\cite{su2024roformer} to arbitrary dimensions by partitioning the model dimension into groups that each encode distinct metadata axes. This design draws on recent extensions of multi-dimensional RoPE in vision language models~\cite{wang2024qwen2}, adapting them to the structure of neural recording configurations. Our ablation confirmed an approximate $3\%$ improvement over standard RoPE, validating the benefit of metadata-aware positional encoding for cross configuration generalization. Given the many sources of potential variability in neural data, even beyond those explored in our work (e.g., neural state variations like alertness/attention, measurement modalities), our results suggest that metadata-aware positional encoding strategies will be important components for neural foundation models and robust decoding architectures. 

Second, neural signals exhibit pronounced non-stationarity due to drift across sessions~\cite{chestek2011long}, yet standard self-attention primarily captures global neural dynamics. Our context-based attention mechanism introduces learnable convolution kernels~\cite{gulati2020conformer} applied to global attention scores to capture local temporal structure, drawing on principles from non-stationary time-series modeling~\cite{liu2022non}. Our ablation analyses revealed a substantial performance gap when replacing context-based attention with standard attention, approximately $5\%$ on cross-task and up to $16\%$ on the cross-site evaluation. This result indicates that the proposed attention mechanism is critical for decoding under distributional shifts present in neural data. 

Third, prior neural transformers employ bidirectional masked autoencoding~\cite{ye2021representation,ye2023neural,ye2025generalist,he2022masked} for offline denoising, which violates the causal structure required for real-time applications. Our causal masked neural modeling enforces unidirectional information flow, transforming the objective from interpolation to extrapolation. Combined with uniform random masking sampled from $\mathcal{U}(0, 1)$, this strategy removes the need for a masking ratio hyperparameter and exposes the model to a range of reconstruction difficulties, yielding more robust representations than any fixed masking configuration. Collectively, these three components form an integrated design that addresses spatial heterogeneity, temporal non-stationarity, and causal representation learning in neural data.

Our work has broad implications for brain-computer interface (BCI) design. Current BCI decoders typically process neural activity in isolation and require frequent recalibration to maintain performance across sessions~\cite{chestek2011long,le2025spint,collinger2013high,silversmith2021plug,gallego2020long,degenhart2020stabilization,natraj2025sampling}. RPNT demonstrated superior cross session and cross site generalization, which shows that self-supervised pretraining on unlabeled neural data can substantially improve decoding robustness~\cite{le2022stndt,zhang2025decoding,ye2023neural}. More robust decoding architectures may help reduce the calibration burden that remains a major obstacle to long-term BCI deployment~\cite{robinson2025application,karpowicz2024few}. Practically, the RPNT pretraining-finetuning paradigm enables a scalable BCI workflow, where a model is pretrained once on large-scale neural recordings and then rapidly adapted to individual users or sessions via FS-SFT with minimal labeled data~\cite{chen2025model}. Such data efficiency is especially relevant in clinical settings where labeled trial collection is time-consuming and burdensome for patients. Moreover, while demonstrated on motor cortex velocity decoding, the RPNT architecture is not restricted to a specific decoded variable or brain region. The framework has natural extensions to broader applications, including closed-loop neuromodulation systems~\cite{fang2021robust,fang2023predictive,fang2024robust} and lightweight neural decoders for resource-constrained clinical settings~\cite{fang2025toward}.

While promising, the limitations of RPNT and our evaluation of it merit further future work. Our modeling and evaluation were restricted to cortical motor areas (PMd and M1), excluding additional cortical and sub-cortical brain regions, such as the frontal cortex and basal ganglia, which contribute to motor control and may have distinct dynamics~\cite{trautmann2025large}. Future work should incorporate recordings from these regions to pretrain more comprehensive neural population models that capture the distributed nature of motor processing. Second, center-out and random target reaching tasks represent relatively simplified and stereotyped motor behaviors.  Natural behaviors, such as continuous object manipulation, sequential movements, and cognitive tasks, involve richer dynamics and feedback interactions with the environment. Validating RPNT on more naturalistic behavioral datasets will be essential to establish broader applicability~\cite{pei2021neural,willett2023high,karpowicz2024few,de2023sharing,safaie2023preserved,banga2025reproducibility}. Third, each RPNT model instance currently processes a single neural spike modality. Brain function involves coordinated dynamics across multiple signal types; local field potentials (LFP), for instance, carry mesoscale oscillatory information complementary to single-unit spiking. Extending RPNT to a multimodal framework that jointly models heterogeneous neural signals ~\cite{wang2024comprehensive,zhang2025neural,caro2023brainlm} represents a key next step toward more comprehensive neural foundation models.

\section*{Methods}
We describe the technical design of RPNT in this section. Briefly speaking, RPNT contains the following three innovative considerations that emerged from the prior investigation of neural transformer architectures~\cite{ye2021representation,ye2023neural,le2025spint,ye2025generalist,ryoo2025generalizable,azabou2023unified}. Figure~\ref{Figure 4: Schematic of the components of RPNT} illustrates the overall architecture design. It includes: 1) Multidimensional rotary positional embedding (MRoPE) that aggregates experimental metadata such as site coordinates, session ids, and behavior types; 2) A context-based attention mechanism that applies convolution kernels to global attention to learn local temporal structure and handle the non-stationarity of neural activity; 3) A robust self supervised learning (SSL) objective with stochastic causal masking strategies and contrastive representation learning. For following each section, we first describe the major difference from the prior work and then the design details.  

\subsection*{MRoPE: Multi-dimensional Rotary Positional Embedding for neural session configurations}
While standard sinusoidal encodings~\cite{vaswani2017attention} and learned positional embeddings~\cite{devlin2019bert} are generally applicable, positional encoding remains challenging for neural data because of signal drift across sessions~\cite{chestek2011long}. Recently, rotary positional embedding (RoPE)~\cite{su2024roformer,azabou2023unified} has become a strong positional encoding approach in pretrained transformer models. Extensions of RoPE to multimodal settings have also shown promise in handling complex positional relationships beyond simple sequential ordering. For example, Qwen2~\cite{wang2024qwen2} demonstrates an effective multi-dimensional RoPE design that captures both sequence position and modality specific structure. Motivated by these developments, we argue that positional encoding for neural data should account for joint recording configurations beyond time alone, so that each neural recording session configuration is represented explicitly through the transformer positional embedding. For example, in recordings across multiple brain sites, an effective positional embedding should model both spatial site location and temporal relationships in order to generalize to unseen site configurations. Thus, we propose a novel extension to RoPE, extending it to multiple dimensions, MRoPE, where $M$ denotes the number of configuration dimensions used to aggregate experimental metadata. This design is more appropriate for neural recordings and fundamentally different from Qwen2, developed for multimodal interactions. 

Using the "Neuro Pixels across Cortical Sites“ dataset (NPCS) as an example, we define MRoPE below. The minimal dimension for this dataset $M$ is $M=3D$, where $D$ stands for the number of electrodes and $3$ corresponds to two dimensional spatial coordinates (denoted by $(x_s, y_s)$) and time $t$. To construct the rotational matrix, we partition the model dimension ($d_{\text{model}}$) into three groups independently ($d_{\text{coord}} = \frac{d_{\text{model}}}{3}$) to represent dimensions each for $x$-coordinate, $y$-coordinate, and temporal position. For each group, we define rotation frequencies:
\begin{equation}
\theta^{(x)}_i = \frac{1}{5000^{2i/d_{\text{coord}}}}, \quad \theta^{(y)}_i = \frac{1}{5000^{2i/d_{\text{coord}}}}, \quad \theta^{(t)}_i = \frac{1}{10000^{2i/d_{\text{coord}}}},
\end{equation}
where $i \in [0, d_{\text{coord}}/2)$. Spatial dimensions use a lower frequency ($f =5000$) to preserve coherence across nearby recording sites, whereas the temporal dimension uses the standard RoPE frequency ($f =10000$) for fine-grained temporal resolution.
\begin{equation}
\label{R_3D}
\mathbf{R}_{3D}^d = \begin{bmatrix}
\mathbf{R}_x(x) & \mathbf{0} & \mathbf{0} \\
\mathbf{0} & \mathbf{R}_y(y) & \mathbf{0} \\
\mathbf{0} & \mathbf{0} & \mathbf{R}_t(t)
\end{bmatrix} \in \mathbb{R}^{d \times d}
\end{equation}
\textbf{Rotation operation.} In contrast to common positional encodings that are added to embeddings, MRoPE, similarly to RoPE in mathematical formulation, operates by applying rotations directly to query and key vectors during attention computation. For a 3D vector that correponds to $M=3$, $(x_s, y_s, t)$, we construct a block-diagonal rotation matrix $\mathbf{R}_\text{3D}(x_s, y_s, t)$(see ~\eqref{R_3D}) that independently rotates each pair of dimensions. The attention score between locations $i$ and $j$ follows $\mathbf{q}_i^T \mathbf{k}_j = \mathbf{q}^T \mathbf{R}_\text{3D}^T(x_i, y_i, t_i) \mathbf{R}_\text{3D}(x_j, y_j, t_j) \mathbf{k} = f(x_j-x_i, y_j-y_i, t_j-t_i)$, which preserves the key property of relative position encoding. This operation supports zero shot generalization to arbitrary brain site configurations while preserving rotational invariance (also see Appendix~\ref{Supplementary: Formulation of MRoPE}).

\begin{figure}
    \centering
    \includegraphics[width=1\linewidth]{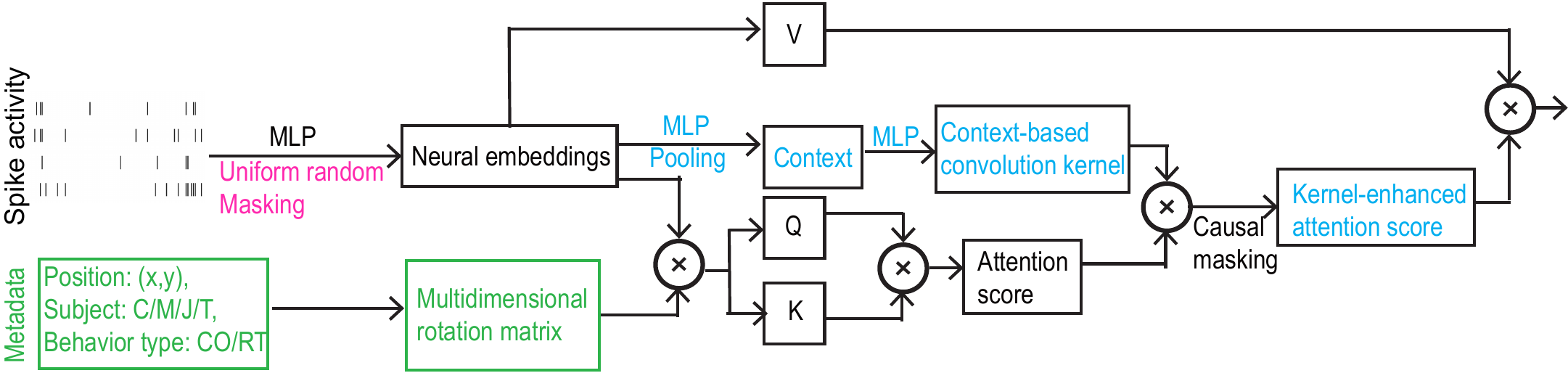}
    \caption{A schematic of the components of RPNT. Components shown in black indicate the standard transformer signal flow, including no masking and a standard attention mechanism. The proposed components are MRoPE (green), context based attention (cyan), and the uniform random masking strategy (pink).\textcolor{black}{MRoPE incorporates experimental metadata into the transformer positional embeddings. The context-based attention captures local temporal structures. The input spike activity is randomly masked according to a uniform distribution for causal masked neural modeling. The masked spike activity is then projected into neural embeddings, forming query, key, and value representations. MRoPE is applied to produce rotary queries and keys. Attention scores are modulated by context based convolutional kernels and combined with causal masking before aggregation with the values.}}
    \label{Figure 4: Schematic of the components of RPNT}
\end{figure}

MRoPE is designed for arbitrary dimensions, allowing it to represent recording configurations that vary across subjects, recording times, locations, and behavior types (see Figure~\ref{Figure 4: Schematic of the components of RPNT}). Specifically, we partition the model dimension $d_{\text{model}}$ into $M$ groups according to the number of the recording configuration dimensions, i.e., $\theta^{(1)}_i = \frac{1}{f_{1}^{2i/d_1}}, \quad \cdots \quad, \theta^{(\mathcal{M})}_i = \frac{1}{f_{M}^{2i/d_{M}}}$, where $d_{1}, \cdots, d_{M}$ are the dimensions of each group and together sum to $d_\text{model}$. The $f_{1}, \cdots, f_{M}$ are the corresponding frequencies. Indeed, for the LTRCH benchmark, we use MRoPE with $M=4D$ to represent behavior type, subject identity, recording time, and temporal position. We used an equal dimension partition ($d_{m} = \frac{d_{model}}{4}$) and set frequencies with 10, 100, 1000, and 10000, respectively (see Appendix~\ref{Supplementary: Formulation of MRoPE} for further details).

\subsection*{Context-based attention mechanism}
Adaptive attention mechanisms have been proposed for domains with nonstationary data, such as video~\cite{chen2021regionvit}. These mechanisms are based on learnable kernels to modulate the attention based on local context. Further, a sliding window attention was introduced to capture local structure~\cite{zhu2021long}, while a hierarchical attention across multiple timescales was employed~\cite{yang2016hierarchical}. Non-stationary Transformers~\cite{liu2022non} introduce series stationary and nonstationary attention to handle distribution shifts in time series forecasting. FEDformer~\cite{zhou2022fedformer} employs frequency-enhanced decomposed architectures to capture both seasonal and trend patterns in nonstationary signals. Crossformer~\cite{zhang2023crossformer} utilizes dimension-segment-wise attention to capture cross-time and cross-dimension dependencies. Motivated by these works and nonstationary nature of neural activity~\cite{chestek2011long}, we introduce a context-based attention mechanism via learnable convolution kernels~\cite{gulati2020conformer} applied to global attention scores to capture local temporal structure. This mechanism incorporates historical context and attention pooling into the attention computation. We also illustrate the attention steps (cyan color) in Figure~\ref{Figure 4: Schematic of the components of RPNT}.

\textbf{Context Generation.} Given a batch neural input $\mathbf{x} \in \mathbb{R}^{B \times T \times D}$ and historical data $\mathbf{H}_{\text{hist}} \in \mathbb{R}^{B \times T_{\text{hist}} \times D}$ containing only past timesteps, we generate a context vector through attention pooling:
\begin{equation}
\mathbf{c} = \text{AttentionPool}(\text{MLP}(\mathbf{H}_{\text{hist}})) \in \mathbb{R}^{B \times D_{\text{context}}}
\end{equation}
We note that $T_{\text{hist}} \subseteq T$. In practice, a simple choice for $T_{\text{hist}}$ is $T_{\text{hist}} = T$, which uses the available historical data to generate the convolution kernels. A more computationally efficient alternative is to use a receding window, $T_{\text{recede}}$, e.g., the most recent $K$ time bins. In this work, we use the first option due to its simplicity.

\textbf{Dynamic kernel generation.} We parameterize the context vector using MLP layers to generate 2D convolution kernels for each attention head
\begin{equation}
\mathbf{K}_h = \text{Reshape}(\text{softmax}(\text{MLP}_{\text{kernel}}(\mathbf{c})), [K_1, K_2]),
\end{equation}
where $h \in [1, H]$ indexes the attention heads and $(K_1, K_2)$ are hyperparameters defining the kernel dimensions.

\textbf{Kernel-enhanced attention.} Finally, we apply a 2D convolution kernel after calculating the standard attention matrix to further capture the local attention structure
\begin{equation}
\mathbf{A}_{\text{kernel}} = \text{Conv2D}(\frac{\mathbf{Q}\mathbf{K}^T}{\sqrt{d}}, \mathbf{K}_h) \odot \mathbf{M}_{\text{causal}},
\end{equation}
where $\mathbf{M}_{\text{causal}}$ is the causal masking ($\mathbf{M}_{\text{causal}}[i,j] = \mathbb{1}[j \leq i]$). This design allows each attention head to learn specialized temporal dependencies for handling neural nonstationarity (see Figure~\ref{Figure 4: Schematic of the components of RPNT}). We follow the convention to apply the causal masking at the very end of the attention operation to preserve the autoregressive property. An alternative causal masking implementation can also be applied within the convolutional operation. In Appendix~\ref{Supplementary: Comparison of two causal masking implementations}, we show that both causal masking implementations make no difference. Thus, we followed the first implementation due to it simplicity.

\subsection*{Neural transformer architecture}
\textbf{Neural encoder.} With the above two components, we construct RPNT's neural transformer encoder, which decouples temporal (for both the NPCS and the LTRCH datasets) and cross site processing (only for the NPCS dataset). First, we build a  \textbf{temporal encoder}. Given neural spike activity $\mathbf{X} \in \mathbb{R}^{B \times S \times T \times N}$ (batch, sites, time, neurons) and corresponding experimental metadata, we first embed each recording site $\mathbf{H}_s^{(0)} = \text{Linear}(\mathbf{X}_{:,s,:,:})$. Then, we build a temporal encoder that consists of $L$ transformer layers with MRoPE and context-based attention  
\begin{equation}
\mathbf{H}_s^{(l+1)} = \mathbf{H}_s^{(l)} + \text{ContextAttn}(\text{LN}(\mathbf{H}_s^{(l)}), \text{MRoPE(configurations)}).
\end{equation}
Most neural representation learning and neural population modeling can be handled by this temporal encoder alone by setting the number of recording sites $S$ to $S=1$, corresponding to repeated neural measurements across sessions. When modeling cross site neural dynamics is needed, such as in sampled Neuropixel recordings for cross site activity (NPCS dataset), we further concatenate a \textbf{spatial encoder} after the temporal encoder. The spatial encoder models cross site interactions through time stepwise processing while naturally preserving causality. At each timestep $t$, we follow the convention to apply the standard multi-head attention (MHA)
\begin{equation}
\mathbf{Z}_{:,:,t,:}^{(l+1)} = \mathbf{Z}_{:,:,t}^{(l)} + \text{MHA}(\text{LN}(\mathbf{Z}_{:,:,t,:}^{(l)})),
\end{equation}
where $\mathbf{Z}^{(0)} = \mathbf{H}^{(L)}$ is taken from the final layer of the temporal encoder. We especially note that the resulting MHA weights can provide interpretable cross-site functional connectivity for NPCS dataset. In our experiments on LTRCH microelectrode benchmark~\cite{perich2018neural}, we focus on demonstrating and validating the temporal encoder. Further, we use the NPCS Neuropixel dataset to provide extra data-driven insights into the MHA map, which discovers the potential brain site interaction.

\textbf{Neural decoder.}
\label{Supplementary: Lightweight decoder}
Our main goal in RPNT is to obtain a pretrained transformer encoder that can later be finetuned for downstream tasks or frozen for data-driven functional connectivity visualization. Accordingly, we design the decoder to be lightweight such that $\text{Decoder}_{\theta}: \mathbb{R}^{B \times T \times D} \rightarrow \mathbb{R}^{B \times T \times N}$ maps encoded representations to Poisson rate parameters through two feedforward blocks and an output projection:
\begin{align}
\mathbf{Z} &= \text{FFN}_2(\text{FFN}_1(\text{LayerNorm}(\mathbf{z}))) \\
\boldsymbol{\lambda} &= \text{Softplus}(\text{Linear}(\mathbf{Z}))
\end{align}

\subsection*{Causal masked neural modeling}
With the neural transformer setup, we next introduce a pretraining framework tailored for neural spike data that robustly performs self-supervised learning (SSL) using masked causal Poisson reconstruction loss with a stochastic masking strategy (uniform random distribution) and contrastive loss for cross-site latent representations.

\textbf{Uniform random masking strategy.}
Unlike fixed masking strategies, which require careful tuning in prior work~\cite{srivastava2014dropout,ye2025generalist,zhang2024towards,he2022masked}, we apply masking in both space (neurons) and time according to a stochastic uniform distribution such that  $p_m \sim \mathcal{U}(0, 1)$, at each batch. This choice is motivated by prior supervised motor decoding studies~\cite{azabou2023unified,le2025spint}. In POYO~\cite{azabou2023unified}, dynamic dropout was shown to improve decoding performance, and in SPINT~\cite{le2025spint}, a uniform neuron sampling strategy further improved supervised motor decoding. These findings motivated us to extend stochastic masking to both temporal and neuronal dimensions during SSL pretraining. This stochastic strategy removes the masking rate hyperparameter while ensuring that the model encounters diverse reconstruction difficulties during pretraining, i.e., $\mathbf{M}_{:,t,n} \sim \text{Bernoulli}(1 - p_m)$.

\textbf{Self-supervised learning objective.} 
Distinct from existing denoising-based transformer work~\cite{ye2021representation}, our approach maintains causality, i.e., reconstruction at time $t$ depends only on unmasked inputs from $t' \leq t$. This adapts the standard masked autoencoding objective~\cite{he2022masked} into a predictive task aligned with the autoregressive nature of neural mechanisms (see discussion section). For masked position $(t, n)$, RPNT reconstructs it with respect to
\begin{equation}
\hat{x}_{:,t,n} = f_\theta(\{\mathbf{x}_{:,t',n'} : (:,t',n') \notin \mathcal{M} \land t' \leq t\}).
\end{equation}
As neural spike counts follow a Poisson distribution, we use the Poisson reconstruction loss
\begin{equation}
\mathcal{L}_{\text{recon}} = \sum_{(:,t,n) \in \mathcal{M}} \left[\hat{\lambda}_{:,t,n} - x_{:,t,n} \log(\hat{\lambda}_{:,t,n} + \epsilon)\right],\\
\end{equation}
where $\hat{\lambda}_{:,t,n} = \text{Decoder}(\mathbf{Z}_{:,t,d})$ and $\epsilon = 10^{-8}$ provides numerical stability. We further include an auxiliary objective to encourage site invariant representations (see Appendix~\ref{Supplementary: Cross-Site Contrastive Learning Details} for details).  To this end, the total combined pretraining loss function $\mathcal{L} = \mathcal{L}_{\text{recon}} + \mu \mathcal{L}_{\text{contrast}}$ balances reconstruction objective with robust site-invariant representation.

\subsection*{Downstream evaluation framework}
We evaluate the generalizability of RPNT on downstream behavior decoding tasks across four scenarios (cross-session, cross-type, cross-subject, and cross-site). Specifically, we build lightweight task-specific heads following the last layer of the pretrained temporal encoder of RPNT
\begin{equation}
\hat{\mathbf{y}}_t = \text{MLP}_{\text{task}}(\mathbf{H}_{:,t}),
\end{equation}
where $\mathbf{H}_{:,t}$ is the temporal encoder outputs across at time $t$. We use mean-squared error (MSE) for SFT on velocity regression tasks and evaluate using $R^2$ metric. We additionally visualize functional connectivity based on the spatial attention maps across time. The spatial encoder's cross site attention weights directly encode functional relationships. We extract the connectivity matrix $\mathbf{C} \in \mathbb{R}^{T \times S \times S}$ by averaging attention weights across layers, i.e., $ \mathbf{C}_{i \rightarrow j}(\text{attention}) = \frac{1}{L} \sum_{l=1}^{L} \text{Attention}_{i \rightarrow j}^{(l)}$. The resulting visualizations are provided in Appendix~\ref{Supplementary: Visualization}.

\section*{Acknowledgements}
The authors acknowledge the partial support of the Harnessing the Data Revolution (HDR) Institute: Accelerated AI Algorithms for Data-Driven Discovery (A3D3), National Science Foundation grant (PHY-2117997 A.L.O., E. S., H.F.). The authors thank the support of the Departments of Electrical and Computer Engineering, Applied Mathematics, and Bioengineering at the University of Washington. For the neuropixel dataset collection, it was supported in part by a National Center for Advancing Translational Sciences of the National Institutes of Health fellowship (TL1 TR002318, R.A.C.), a Nakajima Foundation fellowship (T.O.), a Simons Collaboration for the Global Brain Pilot award (898220, A.L.O.), the National Institute of Neurological Disorders and Stroke (NIH R01 NS134634 and NIH UF1NS126485, A.L.O.), and the NIH Office of Research Infrastructure Programs (P51 OD010425).

\section*{Author contributions statement}
A.L.O. and E.S. supervised the project. H.F., A.L.O., and E.S. conceived the RNPT framework. H.F. implemented the code. HF wrote the initial version of the manuscript. H.F., A.L.O., and E.S. discussed and analyzed the results. H.F., A.L.O., and E.S. reviewed and revised the manuscript. R.A.C, T.O., and B.M. conducted the Neuropixel data collection experiments. All authors reviewed the manuscript before submission.

\section*{Ethics statement}
Our work focuses on algorithm design and evaluation for neural data. To evaluate the RPNT model, we used two datasets: a public dataset and a neuropixels dataset. The public dataset was introduced in the prior work~\cite{perich2018neural} and then made available online~\cite{dandi:DANDI:000688_0.250122.1735}. Data privacy and ethics protocols for experimental procedures related to neural data collection for this data were addressed in the original publication. The neuropixels dataset was collected at the University of Washington and the Washington National Primate Research Center~\cite{canfield2025spatiotemporal}. All procedures were conducted in compliance with the NIH Guide for the Care and Use of Laboratory Animals and were approved by the institutional Animal Care and Use Committee. 

\section*{Reproducibility statement}
All hyperparameters, architectural details, and training configurations necessary for reproducing our results are provided in Appendix~\ref{Appendix: hyperparameters}. Data preprocessing steps and experimental protocols are described in Appendix~\ref{Appendix: preprocessing}. The code will be made publicly available on GitHub upon acceptance of the paper.

\bibliography{reference}

@article{zhang2025decoding,
  title={Decoding inner speech with an end-to-end brain-to-text neural interface},
  author={Zhang, Yizi and others},
  journal={arXiv preprint arXiv:2511.21740},
  year={2025}
}

@article{natraj2025sampling,
  title={Sampling representational plasticity of simple imagined movements across days enables long-term neuroprosthetic control},
  author={Natraj, Nikhilesh and others},
  journal={Cell},
  volume={188},
  number={5},
  pages={1208--1225},
  year={2025},
  publisher={Elsevier}
}

@article{degenhart2020stabilization,
  title={Stabilization of a brain--computer interface via the alignment of low-dimensional spaces of neural activity},
  author={Degenhart, Alan D and others},
  journal={Nature biomedical engineering},
  volume={4},
  number={7},
  pages={672--685},
  year={2020},
  publisher={Nature Publishing Group UK London}
}

@article{gallego2020long,
  title={Long-term stability of cortical population dynamics underlying consistent behavior},
  author={Gallego, Juan A and others},
  journal={Nature neuroscience},
  volume={23},
  number={2},
  pages={260--270},
  year={2020},
  publisher={Nature Publishing Group US New York}
}

@article{silversmith2021plug,
  title={Plug-and-play control of a brain--computer interface through neural map stabilization},
  author={Silversmith, Daniel B and others},
  journal={Nature biotechnology},
  volume={39},
  number={3},
  pages={326--335},
  year={2021},
  publisher={Nature Publishing Group US New York}
}

@article{collinger2013high,
  title={High-performance neuroprosthetic control by an individual with tetraplegia},
  author={Collinger, Jennifer L and others},
  journal={The Lancet},
  volume={381},
  number={9866},
  pages={557--564},
  year={2013},
  publisher={Elsevier}
}

@article{robinson2025application,
  title={An application-based taxonomy for brain--computer interfaces},
  author={Robinson, Jacob T and others},
  journal={Nature Biomedical Engineering},
  volume={9},
  number={6},
  pages={789--791},
  year={2025},
  publisher={Nature Publishing Group UK London}
}

@inproceedings{devlin2019bert,
  title={Bert: Pre-training of deep bidirectional transformers for language understanding},
  author={Devlin, Jacob and Chang, Ming-Wei and Lee, Kenton and Toutanova, Kristina},
  booktitle={Proceedings of the 2019 conference of the North American chapter of the association for computational linguistics: human language technologies, volume 1 (long and short papers)},
  pages={4171--4186},
  year={2019}
}

@article{radford2019language,
  title={Language models are unsupervised multitask learners},
  author={Radford, Alec and Wu, Jeffrey and Child, Rewon and Luan, David and Amodei, Dario and Sutskever, Ilya and others},
  journal={OpenAI blog},
  volume={1},
  number={8},
  pages={9},
  year={2019}
}

@article{dosovitskiy2020image,
  title={An image is worth 16x16 words: Transformers for image recognition at scale},
  author={Dosovitskiy, Alexey and Beyer, Lucas and Kolesnikov, Alexander and Weissenborn, Dirk and Zhai, Xiaohua and Unterthiner, Thomas and Dehghani, Mostafa and Minderer, Matthias and Heigold, Georg and Gelly, Sylvain and others},
  journal={arXiv preprint arXiv:2010.11929},
  year={2020}
}

@article{vaswani2017attention,
  title={Attention is all you need},
  author={Vaswani, Ashish and Shazeer, Noam and Parmar, Niki and Uszkoreit, Jakob and Jones, Llion and Gomez, Aidan N and Kaiser, {\L}ukasz and Polosukhin, Illia},
  journal={Advances in neural information processing systems},
  volume={30},
  year={2017}
}

@book{van2018signal,
  title={Signal processing for neuroscientists},
  author={Van Drongelen, Wim},
  year={2018},
  publisher={Academic press}
}

@article{sani2024dissociative,
  title={Dissociative and prioritized modeling of behaviorally relevant neural dynamics using recurrent neural networks},
  author={Sani, Omid G and Pesaran, Bijan and Shanechi, Maryam M},
  journal={Nature neuroscience},
  volume={27},
  number={10},
  pages={2033--2045},
  year={2024},
  publisher={Nature Publishing Group US New York}
}

@inproceedings{hsieh2025probabilistic,
  title={Probabilistic Geometric Principal Component Analysis with application to neural data},
  author={Hsieh, Han-Lin and Shanechi, Maryam},
  booktitle={The Thirteenth International Conference on Learning Representations},
  year={2025}
}

@article{shanechi2016robust,
  title={Robust brain-machine interface design using optimal feedback control modeling and adaptive point process filtering},
  author={Shanechi, Maryam M and Orsborn, Amy L and Carmena, Jose M},
  journal={PLoS computational biology},
  volume={12},
  number={4},
  pages={e1004730},
  year={2016},
  publisher={Public Library of Science San Francisco, CA USA}
}

@article{orsborn2014closed,
  title={Closed-loop decoder adaptation shapes neural plasticity for skillful neuroprosthetic control},
  author={Orsborn, Amy L and Moorman, Helene G and Overduin, Simon A and Shanechi, Maryam M and Dimitrov, Dragan F and Carmena, Jose M},
  journal={Neuron},
  volume={82},
  number={6},
  pages={1380--1393},
  year={2014},
  publisher={Elsevier}
}

@article{cunningham2014dimensionality,
  title={Dimensionality reduction for large-scale neural recordings},
  author={Cunningham, John P and Yu, Byron M},
  journal={Nature neuroscience},
  volume={17},
  number={11},
  pages={1500--1509},
  year={2014},
  publisher={Nature Publishing Group US New York}
}

@article{semedo2019cortical,
  title={Cortical areas interact through a communication subspace},
  author={Semedo, Jo{\~a}o D and Zandvakili, Amin and Machens, Christian K and Yu, Byron M and Kohn, Adam},
  journal={Neuron},
  volume={102},
  number={1},
  pages={249--259},
  year={2019},
  publisher={Elsevier}
}

@article{yu2008gaussian,
  title={Gaussian-process factor analysis for low-dimensional single-trial analysis of neural population activity},
  author={Yu, Byron M and Cunningham, John P and Santhanam, Gopal and Ryu, Stephen and Shenoy, Krishna V and Sahani, Maneesh},
  journal={Advances in neural information processing systems},
  volume={21},
  year={2008}
}

@article{pandarinath2018inferring,
  title={Inferring single-trial neural population dynamics using sequential auto-encoders},
  author={Pandarinath, Chethan and O’Shea, Daniel J and Collins, Jasmine and Jozefowicz, Rafal and Stavisky, Sergey D and Kao, Jonathan C and Trautmann, Eric M and Kaufman, Matthew T and Ryu, Stephen I and Hochberg, Leigh R and others},
  journal={Nature methods},
  volume={15},
  number={10},
  pages={805--815},
  year={2018},
  publisher={Nature Publishing Group US New York}
}

@article{santhanam2009factor,
  title={Factor-analysis methods for higher-performance neural prostheses},
  author={Santhanam, Gopal and Yu, Byron M and Gilja, Vikash and Ryu, Stephen I and Afshar, Afsheen and Sahani, Maneesh and Shenoy, Krishna V},
  journal={Journal of neurophysiology},
  volume={102},
  number={2},
  pages={1315--1330},
  year={2009},
  publisher={American Physiological Society}
}

@article{ye2021representation,
  title={Representation learning for neural population activity with neural data transformers},
  author={Ye, Joel and Pandarinath, Chethan},
  journal={arXiv preprint arXiv:2108.01210},
  year={2021}
}

@article{ye2023neural,
  title={Neural data transformer 2: multi-context pretraining for neural spiking activity},
  author={Ye, Joel and Collinger, Jennifer and Wehbe, Leila and Gaunt, Robert},
  journal={Advances in Neural Information Processing Systems},
  volume={36},
  pages={80352--80374},
  year={2023}
}

@article{ye2025generalist,
  title={A Generalist Intracortical Motor Decoder},
  author={Ye, Joel and Rizzoglio, Fabio and Smoulder, Adam and Mao, Hongwei and Ma, Xuan and Marino, Patrick and Chowdhury, Raeed and Moore, Dalton and Blumenthal, Gary and Hockeimer, William and others},
  journal={bioRxiv},
  year={2025}
}

@inproceedings{lu2025netformer,
  title={NetFormer: An interpretable model for recovering dynamical connectivity in neuronal population dynamics},
  author={Lu, Ziyu and Zhang, Wuwei and Le, Trung and Wang, Hao and S{\"u}mb{\"u}l, Uygar and SheaBrown, Eric Todd and Mi, Lu},
  booktitle={The Thirteenth International Conference on Learning Representations},
  year={2025}
}

@article{zhang2024towards,
  title={Towards a" universal translator" for neural dynamics at single-cell, single-spike resolution},
  author={Zhang, Yizi and Wang, Yanchen and Jim{\'e}nez-Benet{\'o}, Donato and Wang, Zixuan and Azabou, Mehdi and Richards, Blake and Tung, Renee and Winter, Olivier and Dyer, Eva and Paninski, Liam and others},
  journal={Advances in Neural Information Processing Systems},
  volume={37},
  pages={80495--80521},
  year={2024}
}

@article{ryoo2025generalizable,
  title={Generalizable, real-time neural decoding with hybrid state-space models},
  author={Ryoo, Avery Hee-Woon and Krishna, Nanda H and Mao, Ximeng and Azabou, Mehdi and Dyer, Eva L and Perich, Matthew G and Lajoie, Guillaume},
  journal={arXiv preprint arXiv:2506.05320},
  year={2025}
}

@article{azabou2023unified,
  title={A unified, scalable framework for neural population decoding},
  author={Azabou, Mehdi and Arora, Vinam and Ganesh, Venkataramana and Mao, Ximeng and Nachimuthu, Santosh and Mendelson, Michael and Richards, Blake and Perich, Matthew and Lajoie, Guillaume and Dyer, Eva},
  journal={Advances in Neural Information Processing Systems},
  volume={36},
  pages={44937--44956},
  year={2023}
}

@article{le2022stndt,
  title={Stndt: Modeling neural population activity with spatiotemporal transformers},
  author={Le, Trung and Shlizerman, Eli},
  journal={Advances in Neural Information Processing Systems},
  volume={35},
  pages={17926--17939},
  year={2022}
}

@inproceedings{azabou2024multi,
  title={Multi-session, multi-task neural decoding from distinct cell-types and brain regions},
  author={Azabou, Mehdi and Pan, Krystal Xuejing and Arora, Vinam and Knight, Ian Jarratt and Dyer, Eva L and Richards, Blake Aaron},
  booktitle={The Thirteenth International Conference on Learning Representations},
  year={2024}
}

@article{chau2025population,
  title={Population transformer: Learning population-level representations of neural activity},
  author={Chau, Geeling and Wang, Christopher and Talukder, Sabera and Subramaniam, Vighnesh and Soedarmadji, Saraswati and Yue, Yisong and Katz, Boris and Barbu, Andrei},
  journal={ArXiv},
  pages={arXiv--2406},
  year={2025}
}

@article{wang2023brainbert,
  title={BrainBERT: Self-supervised representation learning for intracranial recordings},
  author={Wang, Christopher and Subramaniam, Vighnesh and Yaari, Adam Uri and Kreiman, Gabriel and Katz, Boris and Cases, Ignacio and Barbu, Andrei},
  journal={arXiv preprint arXiv:2302.14367},
  year={2023}
}

@article{jude2022robust,
  title={Robust alignment of cross-session recordings of neural population activity by behaviour via unsupervised domain adaptation},
  author={Jude, Justin and Perich, Matthew G and Miller, Lee E and Hennig, Matthias H},
  journal={arXiv preprint arXiv:2202.06159},
  year={2022}
}

@article{chestek2011long,
  title={Long-term stability of neural prosthetic control signals from silicon cortical arrays in rhesus macaque motor cortex},
  author={Chestek, Cynthia A and Gilja, Vikash and Nuyujukian, Paul and Foster, Justin D and Fan, Joline M and Kaufman, Matthew T and Churchland, Mark M and Rivera-Alvidrez, Zuley and Cunningham, John P and Ryu, Stephen I and others},
  journal={Journal of neural engineering},
  volume={8},
  number={4},
  pages={045005},
  year={2011},
  publisher={IOP Publishing}
}

@article{su2024roformer,
  title={Roformer: Enhanced transformer with rotary position embedding},
  author={Su, Jianlin and Ahmed, Murtadha and Lu, Yu and Pan, Shengfeng and Bo, Wen and Liu, Yunfeng},
  journal={Neurocomputing},
  volume={568},
  pages={127063},
  year={2024},
  publisher={Elsevier}
}

@inproceedings{he2022masked,
  title={Masked autoencoders are scalable vision learners},
  author={He, Kaiming and Chen, Xinlei and Xie, Saining and Li, Yanghao and Doll{\'a}r, Piotr and Girshick, Ross},
  booktitle={Proceedings of the IEEE/CVF conference on computer vision and pattern recognition},
  pages={16000--16009},
  year={2022}
}

@article{perich2018neural,
  title={A neural population mechanism for rapid learning},
  author={Perich, Matthew G and Gallego, Juan A and Miller, Lee E},
  journal={Neuron},
  volume={100},
  number={4},
  pages={964--976},
  year={2018},
  publisher={Elsevier}
}

@article{le2025spint,
  title={SPINT: Spatial Permutation-Invariant Neural Transformer for Consistent Intracortical Motor Decoding},
  author={Le, Trung and Fang, Hao and Li, Jingyuan and Nguyen, Tung and Mi, Lu and Orsborn, Amy and S{\"u}mb{\"u}l, Uygar and Shlizerman, Eli},
  journal={arXiv preprint arXiv:2507.08402},
  year={2025}
}

@article{chen2021regionvit,
  title={Regionvit: Regional-to-local attention for vision transformers},
  author={Chen, Chun-Fu and Panda, Rameswar and Fan, Quanfu},
  journal={arXiv preprint arXiv:2106.02689},
  year={2021}
}

@article{zhu2021long,
  title={Long-short transformer: Efficient transformers for language and vision},
  author={Zhu, Chen and Ping, Wei and Xiao, Chaowei and Shoeybi, Mohammad and Goldstein, Tom and Anandkumar, Anima and Catanzaro, Bryan},
  journal={Advances in neural information processing systems},
  volume={34},
  pages={17723--17736},
  year={2021}
}

@inproceedings{yang2016hierarchical,
  title={Hierarchical attention networks for document classification},
  author={Yang, Zichao and Yang, Diyi and Dyer, Chris and He, Xiaodong and Smola, Alex and Hovy, Eduard},
  booktitle={Proceedings of the 2016 conference of the North American chapter of the association for computational linguistics: human language technologies},
  pages={1480--1489},
  year={2016}
}

@article{glaser2020machine,
  title={Machine learning for neural decoding},
  author={Glaser, Joshua I and Benjamin, Ari S and Chowdhury, Raeed H and Perich, Matthew G and Miller, Lee E and Kording, Konrad P},
  journal={eneuro},
  volume={7},
  number={4},
  year={2020},
  publisher={Society for Neuroscience}
}

@article{cho2014learning,
  title={Learning phrase representations using RNN encoder-decoder for statistical machine translation},
  author={Cho, Kyunghyun and Van Merri{\"e}nboer, Bart and Gulcehre, Caglar and Bahdanau, Dzmitry and Bougares, Fethi and Schwenk, Holger and Bengio, Yoshua},
  journal={arXiv preprint arXiv:1406.1078},
  year={2014}
}

@article{gu2021efficiently,
  title={Efficiently modeling long sequences with structured state spaces},
  author={Gu, Albert and Goel, Karan and R{\'e}, Christopher},
  journal={arXiv preprint arXiv:2111.00396},
  year={2021}
}

@article{gu2023mamba,
  title={Mamba: Linear-time sequence modeling with selective state spaces},
  author={Gu, Albert and Dao, Tri},
  journal={arXiv preprint arXiv:2312.00752},
  year={2023}
}

@article{trautmann2025large,
  title={Large-scale high-density brain-wide neural recording in nonhuman primates},
  author={Trautmann, Eric M and Hesse, Janis K and Stine, Gabriel M and Xia, Ruobing and Zhu, Shude and O’Shea, Daniel J and Karsh, Bill and Colonell, Jennifer and Lanfranchi, Frank F and Vyas, Saurabh and others},
  journal={Nature Neuroscience},
  pages={1--14},
  year={2025},
  publisher={Nature Publishing Group US New York}
}

@article{willett2023high,
  title={A high-performance speech neuroprosthesis},
  author={Willett, Francis R and Kunz, Erin M and Fan, Chaofei and Avansino, Donald T and Wilson, Guy H and Choi, Eun Young and Kamdar, Foram and Glasser, Matthew F and Hochberg, Leigh R and Druckmann, Shaul and others},
  journal={Nature},
  volume={620},
  number={7976},
  pages={1031--1036},
  year={2023},
  publisher={Nature Publishing Group UK London}
}

@article{wu2006bayesian,
  title={Bayesian population decoding of motor cortical activity using a Kalman filter},
  author={Wu, Wei and Gao, Yun and Bienenstock, Elie and Donoghue, John P and Black, Michael J},
  journal={Neural computation},
  volume={18},
  number={1},
  pages={80--118},
  year={2006},
  publisher={MIT Press One Rogers Street, Cambridge, MA 02142-1209, USA journals-info~…}
}

@article{karpowicz2024few,
  title={Few-shot algorithms for consistent neural decoding (falcon) benchmark},
  author={Karpowicz, Brianna and Ye, Joel and Fan, Chaofei and Tostado-Marcos, Pablo and Rizzoglio, Fabio and Washington, Clayton and Scodeler, Thiago and de Lucena, Diogo and Nason-Tomaszewski, Samuel and Mender, Matthew and others},
  journal={Advances in Neural Information Processing Systems},
  volume={37},
  pages={76578--76615},
  year={2024}
}

@article{pei2021neural,
  title={Neural Latents Benchmark'21: Evaluating latent variable models of neural population activity},
  author={Pei, Felix and Ye, Joel and Zoltowski, David and Wu, Anqi and Chowdhury, Raeed H and Sohn, Hansem and O'Doherty, Joseph E and Shenoy, Krishna V and Kaufman, Matthew T and Churchland, Mark and others},
  journal={arXiv preprint arXiv:2109.04463},
  year={2021}
}

@article{wang2024comprehensive,
  title={A comprehensive review of multimodal large language models: Performance and challenges across different tasks},
  author={Wang, Jiaqi and Jiang, Hanqi and Liu, Yiheng and Ma, Chong and Zhang, Xu and Pan, Yi and Liu, Mengyuan and Gu, Peiran and Xia, Sichen and Li, Wenjun and others},
  journal={arXiv preprint arXiv:2408.01319},
  year={2024}
}

@article{srivastava2014dropout,
  title={Dropout: a simple way to prevent neural networks from overfitting},
  author={Srivastava, Nitish and Hinton, Geoffrey and Krizhevsky, Alex and Sutskever, Ilya and Salakhutdinov, Ruslan},
  journal={The journal of machine learning research},
  volume={15},
  number={1},
  pages={1929--1958},
  year={2014},
  publisher={JMLR. org}
}

@article{wu2002neural,
  title={Neural decoding of cursor motion using a Kalman filter},
  author={Wu, Wei and Black, M and Gao, Yun and Serruya, M and Shaikhouni, A and Donoghue, J and Bienenstock, Elie},
  journal={Advances in neural information processing systems},
  volume={15},
  year={2002}
}

@article{mante2013context,
  title={Context-dependent computation by recurrent dynamics in prefrontal cortex},
  author={Mante, Valerio and Sussillo, David and Shenoy, Krishna V and Newsome, William T},
  journal={nature},
  volume={503},
  number={7474},
  pages={78--84},
  year={2013},
  publisher={Nature Publishing Group UK London}
}

@article{sussillo2016making,
  title={Making brain--machine interfaces robust to future neural variability},
  author={Sussillo, David and Stavisky, Sergey D and Kao, Jonathan C and Ryu, Stephen I and Shenoy, Krishna V},
  journal={Nature communications},
  volume={7},
  number={1},
  pages={13749},
  year={2016},
  publisher={Nature Publishing Group UK London}
}

@article{fang2025toward,
  title={Toward Lightweight and Fast Inference Neural Decoder Design using Quantization-Aware Training: A Simulation Study},
  author={Fang, Hao and others},
  journal={Authorea Preprints},
  year={2025},
  publisher={Authorea}
}

@inproceedings{fang2021robust,
  title = {A Robust and Adaptive Control Algorithm for Closed-Loop Brain Stimulation},
  author = {Fang, Hao and Yang, Yuxiao},
  booktitle = {2021 43rd Annual International Conference of the IEEE Engineering in Medicine \& Biology Society (EMBC)},
  pages = {6049--6052},
  year = {2021},
  organization = {IEEE}
}

@article{fang2024robust,
  title={Robust adaptive deep brain stimulation control of in-silico non-stationary Parkinsonian neural oscillatory dynamics},
  author={Fang, Hao and others},
  journal={J. Neural Eng.},
  volume={21},
  number={3},
  pages={036043},
  year={2024},
  publisher={IOP Publishing}
}

@article{fang2023predictive,
  title={Predictive neuromodulation of cingulo-frontal neural dynamics in major depressive disorder using a brain-computer interface system: A simulation study},
  author={Fang, Hao and Yang, Yuxiao},
  journal={Front. comput. neurosci},
  volume={17},
  pages={1119685},
  year={2023},
  publisher={Frontiers Media SA}
}

@article{chen2025model,
  title={Model-agnostic meta-learning for EEG-based inter-subject emotion recognition},
  author={Chen, Cheng and others},
  journal={J. Neural Eng.},
  volume={22},
  number={1},
  pages={016008},
  year={2025},
  publisher={IOP Publishing}
}

@article{canfield2025spatiotemporal,
  title={The spatiotemporal structure of neural activity in motor cortex during reaching},
  author={Canfield, Ryan A and Ouchi, Tomohiro and Fang, Hao and Macagno, Beatrice and Smith, Lydia I and Scholl, Leo R and Orsborn, Amy L},
  journal={bioRxiv},
  pages={2025--10},
  year={2025},
  publisher={Cold Spring Harbor Laboratory}
}

@article{liu2022non,
  title={Non-stationary transformers: Exploring the stationarity in time series forecasting},
  author={Liu, Yong and Wu, Haixu and Wang, Jianmin and Long, Mingsheng},
  journal={Advances in neural information processing systems},
  volume={35},
  pages={9881--9893},
  year={2022}
}

@inproceedings{zhou2022fedformer,
  title={Fedformer: Frequency enhanced decomposed transformer for long-term series forecasting},
  author={Zhou, Tian and Ma, Ziqing and Wen, Qingsong and Wang, Xue and Sun, Liang and Jin, Rong},
  booktitle={International conference on machine learning},
  pages={27268--27286},
  year={2022},
  organization={PMLR}
}

@inproceedings{zhang2023crossformer,
  title={Crossformer: Transformer utilizing cross-dimension dependency for multivariate time series forecasting},
  author={Zhang, Yunhao and Yan, Junchi},
  booktitle={The eleventh international conference on learning representations},
  year={2023}
}

@article{wang2024qwen2,
  title={Qwen2-vl: Enhancing vision-language model's perception of the world at any resolution},
  author={Wang, Peng and Bai, Shuai and Tan, Sinan and Wang, Shijie and Fan, Zhihao and Bai, Jinze and Chen, Keqin and Liu, Xuejing and Wang, Jialin and Ge, Wenbin and others},
  journal={arXiv preprint arXiv:2409.12191},
  year={2024}
}

@article{gulati2020conformer,
  title={Conformer: Convolution-augmented transformer for speech recognition},
  author={Gulati, Anmol and Qin, James and Chiu, Chung-Cheng and Parmar, Niki and Zhang, Yu and Yu, Jiahui and Han, Wei and Wang, Shibo and Zhang, Zhengdong and Wu, Yonghui and others},
  journal={arXiv preprint arXiv:2005.08100},
  year={2020}
}

@article{de2023sharing,
  title={Sharing neurophysiology data from the Allen Brain Observatory},
  author={de Vries, Saskia EJ and Siegle, Joshua H and Koch, Christof},
  journal={Elife},
  volume={12},
  pages={e85550},
  year={2023},
  publisher={eLife Sciences Publications Limited}
}

@article{safaie2023preserved,
  title={Preserved neural dynamics across animals performing similar behaviour},
  author={Safaie, Mostafa and Chang, Joanna C and Park, Junchol and Miller, Lee E and Dudman, Joshua T and Perich, Matthew G and Gallego, Juan A},
  journal={Nature},
  volume={623},
  number={7988},
  pages={765--771},
  year={2023},
  publisher={Nature Publishing Group UK London}
}

@article{banga2025reproducibility,
  title={Reproducibility of in vivo electrophysiological measurements in mice},
  author={Banga, Kush and Benson, Julius and Bhagat, Jai and Biderman, Dan and Birman, Daniel and Bonacchi, Niccol{\`o} and Bruijns, Sebastian A and Buchanan, Kelly and Campbell, Robert AA and Carandini, Matteo and others},
  journal={Elife},
  volume={13},
  pages={RP100840},
  year={2025},
  publisher={eLife Sciences Publications Limited}
}

@article{zhang2025neural,
  title={Neural encoding and decoding at scale},
  author={Zhang, Yizi and Wang, Yanchen and Azabou, Mehdi and Andre, Alexandre and Wang, Zixuan and Lyu, Hanrui and Laboratory, The International Brain and Dyer, Eva and Paninski, Liam and Hurwitz, Cole},
  journal={arXiv preprint arXiv:2504.08201},
  year={2025}
}

@article{caro2023brainlm,
  title={BrainLM: A foundation model for brain activity recordings},
  author={Caro, Josue Ortega and Fonseca, Antonio H de O and Averill, Christopher and Rizvi, Syed A and Rosati, Matteo and Cross, James L and Mittal, Prateek and Zappala, Emanuele and Levine, Daniel and Dhodapkar, Rahul M and others},
  journal={BioRxiv},
  pages={2023--09},
  year={2023},
  publisher={Cold Spring Harbor Laboratory}
}

@dataset{dandi:DANDI:000688_0.250122.1735,
  title = {Long-term recordings of motor and premotor cortical spiking activity during reaching in monkeys},
  author = {Perich, Matthew G. and Miller, Lee E. and Azabou, Mehdi and Dyer, Eva L.},
  doi = {10.48324/dandi.000688/0.250122.1735},
  url = {https://dandiarchive.org/dandiset/000688/0.250122.1735},
  version = {0.250122.1735},
  publisher = {DANDI Archive},
  note = {DANDI:DANDI:000688/0.250122.1735}
}

\section*{Appendix}
\setcounter{section}{0}
\setcounter{equation}{0}
\setcounter{table}{0}
\setcounter{figure}{0}

\section{Micro-electrode Data Pre-processing Pipeline}
\label{Supplementary: Data Pre-processing Pipeline}
We design a custom preprocessing pipeline to construct the training dataset for pretraining and the test dataset for downstream evaluation via \textit{torchbrain} for the LTRCH benchmark~\cite{perich2018neural,dandi:DANDI:000688_0.250122.1735}. In particular, we create a SpikeBinner module to convert spike timestamps into binned spike counts for transformer input, where each sample spans 1s using 20ms bins without overlap, resulting in tensors of shape (windows, 50, units). For the pretraining dataset, we filter the data from $intervals.start$ to $intervals.end$, and thus the data is not trial-aligned. For the downstream evaluation dataset, we use data from $intervals.go\ cue\ time$ to $intervals.stoptime$ for center out tasks, and again use data from $intervals.start$ to $intervals.end$ for random target tasks. We construct the data using non-overlapping sequential bins within each session; we do not randomly shuffle the dataset to respect the temporal order. Finally, we use the same temporal split of $70\%$, $10\%$, $20\%$ for train/val/test splits as in POSSM~\cite{ryoo2025generalizable}.

To batch the data efficiently for transformer training, we adopt the following strategy to deal with a varying number of recorded neurons across sessions. Specifically, RPNT randomly resamples the neural activity to a fixed maximum number of neurons. For example, for a session with data of shape $(T, 48)$, where 48 is the number of neurons, we resample the data to match the batch shape $(T, 100)$. In this case, we randomly sample an additional 52 neuron traces to reach the target size. For instance, the $(T, 48)$ data may be repeated once and then supplemented with the first four neurons $(T, 4)$ to obtain $(T, 52)$, which is then concatenated with the original data to form $(T, 100)$. The same strategy applies to sessions with more than 100 neurons, in which case 100 neurons are randomly selected to match the fixed shape $(T, 100)$. In practice, this simple resampling strategy works well because stochastic uniform spatial and temporal masking is always applied during RPNT pretraining. In our experiments, we set the fixed number of neurons to $300$.

\section{Neuropixel Data Preprocessing}
\label{Appendix: preprocessing}

We require batch-like neural data for transformer model pretraining. However, raw data collected across heterogeneous recording sites vary in trial count, neuron population, and recording duration. Let $\mathcal{S} = \{S_1, S_2, \ldots, S_K\}$ denote a collection of $K$ recording sites, where each site $S_i$ contains neural spike data $\mathbf{X}_i \in \mathbb{R}^{C_i \times T_i \times N_i}$ with $C_i$ successful trials, $t_j$ time points, and $N_i$ recording neurons after spike sorting. Our objective is to construct a unified dataset $\mathcal{D} = \{(\mathbf{X}_{k=1}^K), \mathbf{L}\}$ where $\mathbf{X}^{(k)} \in \mathbb{R}^{C \times T \times N}$ denotes standardized neural activity for site and $\mathbf{L} \in \mathbb{R}^{L \times 2}$ encodes spatial location coordinates.

Therefore, we implement a systematic pipeline to load multi site Neuropixel recordings from the NPCS benchmark. Given a metadata file containing site locations $\mathbf{L}_i = (x_i, y_i) \in \mathbb{R}^2$ and the corresponding data files, we filter sites based on data availability and quality criteria. Each site's spike data is loaded as $\mathbf{X}_i \in \mathbb{R}^{C_i \times T_i \times N_i}$. We first extract the 1s reaching period data and bin the data using a $20ms$ bin size. This process yields a uniform temporal dimension ($T=50$). Next, we standardize neuron numbers across heterogeneous sites while preserving variability through neuron subset sampling. For each train/valid/test split $s$ (based on trials) and site $i$ with shape $\mathbf{X}_i^{(s)} \in \mathbb{R}^{C_i^{(s)} \times T \times N_i}$, we apply the pipeline in algorithm~\ref{alg:preprocessing}.

\begin{algorithm}[h]
\caption{Data Preprocessing Pipeline on the Neuropixel Dataset}
\label{alg:preprocessing}
\begin{algorithmic}[1]
\REQUIRE Site data $\mathbf{X}_i \in \mathbb{R}^{C_i \times T \times N_i}$, target neurons $N$, sample times $M$, target trial counts $C_{\text{target}}^{(s)}$
\ENSURE Standardized data $\hat{\mathbf{X}}_i^{(s)} \in \mathbb{R}^{T_{\text{target}}^{(s)} \times T \times N}$

\STATE \textbf{Step 1: Split} - Apply train/val/test split (e.g., $80\%/10\%/10\%$) to get $\mathbf{X}_i^{(s)}$

\STATE \textbf{Step 2: Neuron Multi-Sampling}
\STATE $N_{\text{total}} \leftarrow N \cdot M$
\IF{$N_i \geq N_{\text{total}}$}
    \STATE $\boldsymbol{\nu} \leftarrow \text{sample\_without\_replacement}(N_i, N_{\text{total}})$
\ELSE
    \STATE $\boldsymbol{\nu} \leftarrow \text{sample\_with\_replacement}(N_i, N_{\text{total}})$
\ENDIF
\STATE $\mathbf{Y} \leftarrow \mathbf{X}_i^{(s)}[:, :, \boldsymbol{\nu}]$ \COMMENT{Neuron sampling}
\STATE $\mathbf{Z} \leftarrow \text{reshape}(\mathbf{Y}, (C_i^{(s)} \cdot M, T, N))$ \COMMENT{Multi-sampling}

\STATE \textbf{Step 3: Trial Sampling for Target Matching}
\STATE $T_{\text{available}} \leftarrow T_i^{(s)} \cdot M$
\IF{$T_{\text{available}} \geq T_{\text{target}}^{(s)}$}
    \STATE $\boldsymbol{\tau} \leftarrow \text{sample\_without\_replacement}(C_{\text{available}}, C_{\text{target}}^{(s)})$
\ELSE
    \STATE $\boldsymbol{\tau} \leftarrow \text{sample\_with\_replacement}(C_{\text{available}}, C_{\text{target}}^{(s)})$
\ENDIF
\STATE $\hat{\mathbf{X}}_i^{(s)} \leftarrow \mathbf{Z}[\boldsymbol{\tau}, :, :]$ \COMMENT{Final trial sampling}
\end{algorithmic}
\end{algorithm}

After neuron multi-sampling, we apply a split-specific trial sampling to match target trial counts
\begin{equation}
C_{\text{target}}^{(s)} = \begin{cases}
C_{\text{train}} & \text{if } s = \text{train} \\
C_{\text{min}} & \text{if } s \in \{\text{val}, \text{test}\},
\end{cases}
\end{equation}
where $T_{\text{train}}$ is the target training sample count and $T_{\text{min}}$ is the minimal validation/test count to prevent oversampling. Finally, we aggregate the standardized site data into  tensors with consistent dimensions
\begin{align}
\mathbf{X}_{\text{sites}}^{(s)} &= \text{stack}([\hat{\mathbf{X}}_1^{(s)}, \hat{\mathbf{X}}_2^{(s)}, \ldots, \hat{\mathbf{X}}_K^{(s)}]) \in \mathbb{R}^{K \times C_{\text{target}}^{(s)} \times T \times N}, \\
\mathbf{X}^{(s)} &= \text{transpose}(\mathbf{X}_{\text{sites}}^{(s)}, (1, 0, 2, 3)) \in \mathbb{R}^{C_{\text{target}}^{(s)} \times K \times T \times N}.
\end{align}
Our data processing ensures strict separation between between training, validation, and test sets ($\mathbf{X}_i^{\text{train}} \cap \mathbf{X}_i^{\text{val}} \cap \mathbf{X}_i^{\text{test}} = \emptyset, \quad \forall i \in \{1, \ldots, K\}$) while maintaining consistent tensor dimensionality, i.e., $\mathbf{X}^{(s)} \in \mathbb{R}^{B^{(s)} \times K \times C \times N}, \quad \forall s \in \{\text{train}, \text{val}, \text{test}\}$, where $B^{(s)} = \tilde{C}^{(s)}$ denotes the batch dimension for split $s$.

Raw trial counts range from 200 to 1417 per site, and raw neuron counts range from 83 to 703 per site. Our preprocessing pipeline standardizes all site data, yielding 80,000 (16*5000) training samples for RPNT pretraining together with corresponding validation and test sets. The resulting dataset is agnostic to neuron count and neuron ordering; instead, it depends only on the target number of neurons $N$ specified by the user. As a result, RPNT trained on this dataset can accommodate arbitrary neuron counts through the sampling strategy implemented in the pipeline. To this end, Tables~\ref{table: site_breakdown} and Table~\ref{table: dataset_stats} summarize the pretraining data statistics before and after preprocessing.

\begin{table}
\centering
\begin{tabular}{l|c|c|c|c|c}
\hline
\textbf{Site ID} & \textbf{Coordinates} & \textbf{Trials} & \textbf{Time bins} & \textbf{Neurons} & \textbf{Site labels}\\
\hline
9940 & (-1, 1) & 1286 & 50 & 328 & S13\\
\hline
10802 & (3, 2) & 1200 & 50 & 703 & S11\\
\hline
10812 & (-3, 2) & 1200 & 50 & 511 & S9\\
\hline
10820 & (-2, 4) & 1086 & 50 & 394 & S3\\
\hline
10828 & (-1, 4) & 1417 & 50 & 329 & S4\\
\hline
12269 & (3, -4) & 752 & 50 & 252 & S16\\
\hline
12290 & (5, 2) & 763 & 50 & 367 & S12\\
\hline
13153 & (1, 5) & 415 & 50 & 275 & S2\\
\hline
13122 & (2, 4) & 674 & 50 & 83 & S6\\
\hline
13239 & (-1, 5) & 967 & 50 & 154 & S1\\
\hline
13256 & (2, -4) & 1024 & 50 & 255 & S15\\
\hline
13272 & (-1, 3) & 1056 & 50 & 187 & S7\\
\hline
14116 (Test) & (3, -2) & 789 & 50 & 258 & S17\\
\hline
14139 & (2, 2) & 865 & 50 & 125 & S10\\
\hline
14824 & (2, 3) & 200 & 50 & 354 & S8\\
\hline
14878 & (3, -3) & 200 & 50 & 309 & S14\\
\hline
14891 & (0, 4) & 200 & 50 & 106 & S5\\
\hline
\end{tabular}
\caption{Breakdown of data characteristics by site.}
\label{table: site_breakdown}

\vspace{0.5cm}

\centering
\begin{tabular}{|l|c|c|c|}
\hline
\textbf{Split} & \textbf{Samples} & \textbf{Shape} & \textbf{Sampling Strategy} \\
\hline
Training & 80,000 & $(5000, 16, 50, 300)$ & Full sampling \\
\hline
Validation & 3,200 & $(200, 16, 50, 300)$ & Minimal sampling \\
\hline
Test & 3,200 & $(200, 16, 50, 300)$ & Minimal sampling \\
\hline
\end{tabular}
\caption{Characteristics of dataset splits after preprocessing. Format: (Trials, Sites, Time, Neurons).}
\label{table: dataset_stats}
\end{table}

\section{Formulation of MRoPE}
\label{Supplementary: Formulation of MRoPE}
We first use 3D-RoPE as a concrete illustration of the general MRoPE formulation, and we demonstrate the extension to 4D-RoPE later. For neural recordings from $S$ sites over $T$ timesteps, we require positional encodings $\mathbf{PE}(s, t, d) \in \mathbb{R}^{d_{model}}$ where $d$ is the model embedding dimension. A desirable positional encoding should satisfy two key properties: (1)~\textit{site specific}, such that $\mathbf{PE}(s_i, t, d) \neq \mathbf{PE}(s_j, t, d)$ for $i \neq j$, ensuring distinct representations for different site locations (2)~\textit{zero-shot generalization}, such that the the encoding function can generate reasonable embeddings for previously unseen sites $\mathbf{s}_{new} = (x_{new}, y_{new})$.
\subsection*{Extension from standard RoPE to 3D-RoPE}
The key insight of standard RoPE is that attention scores depend only on relative distances. By combining RoPE with session configuration variables, namely spatial coordinates $(x, y)$ for site $s$ and temporal position $t$, our goal is to construct 3D-RoPE such that the attention score depends on relative spatial and temporal distances $f(|x_i-x_j|, |y_i-y_j|, |t_i-t_j|)$. For position $(x, y, t)$, the complete 3D-RoPE transformation is a block-diagonal matrix
\begin{equation}
\mathbf{R}_{3D}^d = \begin{bmatrix}
\mathbf{R}_x(x) & \mathbf{0} & \mathbf{0} \\
\mathbf{0} & \mathbf{R}_y(y) & \mathbf{0} \\
\mathbf{0} & \mathbf{0} & \mathbf{R}_t(t)
\end{bmatrix} \in \mathbb{R}^{d \times d}
\end{equation}
where each sub-matrix $\mathbf{R}_{m \in{\{x,y,t\}}}$ consists of 2×2 rotation blocks:
\begin{equation}
\label{each entry representation}
\mathbf{R}_m(m) = \begin{bmatrix}
\cos(m\theta^{(m}_0) & -\sin(m\theta^{(m)}_0) & & \\
\sin(m\theta^{(m)}_0) & \cos(m\theta^{(m)}_0) & & \\
& & \ddots & \\
& & & \cos(m\theta^{(m)}_{\frac{d}{6}-1}) & -\sin(m\theta^{(m)}_{\frac{d}{6}-1}) \\
& & & \sin(m\theta^{(m)}_{\frac{d}{6}-1}) & \cos(m\theta^{(m)}_{\frac{d}{6}-1})
\end{bmatrix}.
\end{equation}

\subsection*{Relative Position Property Proof}
3D-RoPE maintains the RoPE property that the attention scores depend only on relative configuration. For two configureations $(x_i, y_i, t_i)$ and $(x_j, y_j, t_j)$:
\begin{align}
\mathbf{q}_i^T \mathbf{k}_j &= (\mathbf{R}_{3D}(x_i, y_i, t_i)\mathbf{q})^T (\mathbf{R}_{3D}(x_j, y_j, t_j)\mathbf{k}) \\
&= \mathbf{q}^T \mathbf{R}_{3D}^T(x_i, y_i, t_i) \mathbf{R}_{3D}(x_j, y_j, t_j) \mathbf{k} \\
&= \mathbf{q}^T \mathbf{R}_{3D}(x_j-x_i, y_j-y_i, t_j-t_i) \mathbf{k}
\end{align}
This follows from the rotation composition property: $\mathbf{R}^T(\alpha)\mathbf{R}(\beta) = \mathbf{R}(\beta-\alpha)$. This design ensures \textbf{site-specificity} since different spatial coordinates produce different rotations, and enables \textbf{zero-shot generalization} as the rotation operations naturally handle arbitrary continuous spatial coordinates.

\subsection*{Other positional embedding baselines}
\textbf{Original RoPE.}
\begin{align}
\mathbf{R}^d_{\Theta,m} = \begin{bmatrix}
\cos m\theta_0 & -\sin m\theta_0 & 0 & 0 & \cdots & 0 & 0 \\
\sin m\theta_0 & \cos m\theta_0 & 0 & 0 & \cdots & 0 & 0 \\
0 & 0 & \cos m\theta_1 & -\sin m\theta_1 & \cdots & 0 & 0 \\
0 & 0 & \sin m\theta_1 & \cos m\theta_1 & \cdots & 0 & 0 \\
\vdots & \vdots & \vdots & \vdots & \ddots & \vdots & \vdots \\
0 & 0 & 0 & 0 & \cdots & \cos m\theta_{d/2-1} & -\sin m\theta_{d/2-1} \\
0 & 0 & 0 & 0 & \cdots & \sin m\theta_{d/2-1} & \cos m\theta_{d/2-1}
\end{bmatrix}
\end{align}

\textbf{Learnable PE.} We implement an improved version of the learnable PE~\cite{devlin2019bert} so that it has a simple mechanism for the configuration generalization. Thus, we employ an MLP that directly maps spatial-temporal coordinates to the embedding space:
\begin{align}
\mathbf{PE}(s, t, d) = \text{MLP}_{pos}([x \cdot \alpha, y \cdot \alpha, t])
\end{align}
where $\alpha$ is a spatial scaling factor. In our experiments, we set the $\alpha = 1$ for simplicity. \textbf{Sinusoidal PE.} we also incorporate the standard sinusoidal positional embedding~\cite{vaswani2017attention} in our ablation study.

\subsection*{4D-RoPE in the public benchmark}
\label{section: 4D-RoPE in the public benchmark}
We showed the following rotation matrix for $M=4D$ that was used on the public benchmark
\begin{equation}
\mathbf{R}_{4D}^d = \begin{bmatrix}
\mathbf{R}_\text{task}(x) & \mathbf{0} & \mathbf{0} & \mathbf{0}\\
\mathbf{0} & \mathbf{R}_\text{subject}(y) & \mathbf{0} & \mathbf{0}\\
\mathbf{0} & \mathbf{0} & \mathbf{R}_\text{recording time}(t) & \mathbf{0} \\
\mathbf{0} & \mathbf{0} & \mathbf{0} & \mathbf{R}_t(t) & 
\end{bmatrix} \in \mathbb{R}^{d \times d}
\end{equation}
where each entry was constructed similarly to ~\eqref{each entry representation}. Different from the Neuropixel datasets, in which the $(x,y)$ coordinates have values, the public benchmark was string-based metadata. Therefore, we used the discrete embeddings for representation. Specifically, we used i) $[0, 1]$ to embed behavior types $\{\text{CO}, \text{RT}\}$; ii) $[0, 1, 2, 3]$ to embed subject $\{c, j, m, t\}$; iii) $[0,1]$ range normalization to the recording time (day/month/year). The choice of (i) is according to alphabetical order. To show that the alphabetical discrete embedding does not introduce inductive bias. We further performed another experiment by randomly changing the dataset assigned numbers to $[1,3,2,0]$ for $\{c, j, m, t\}$, and tested it on the T-RT test.  Table~\ref{table: Sweep study on subject embeddings} did not show an obvious performance gap in different subject embeddings. 
\begin{table}
\centering
\begin{tabular}{l|c}
\hline
Discrete subject embedding & T-RT \\
\hline 
$[1,3,2,0]$ & 0.8479 $\pm$ 0.0960 \\
$[0,1,2,3]$ & 0.8515 $\pm$ 0.1071\\
\hline
\end{tabular}
\caption{Sweep study on subject embeddings}
\label{table: Sweep study on subject embeddings}
\end{table}
We further envision that using learnable MLPs for the string-based meta information may further improve the embedding representation. However, this is more about the incremental improvement of our MRoPE, and thus will not be investigated here. 

\section{Pretraining Details}

\subsection*{Cross-Site Contrastive Learning Details}
\label{Supplementary: Cross-Site Contrastive Learning Details}
\begin{table}
\centering
\begin{tabular}{l|c}
\hline
Method & Cross-Site (B-CS) \\
\hline 
w.o contrastive loss & 0.6267 $\pm$ 0.0277 \\
\textbf{w. contrastive loss} & \textbf{0.6612} $\pm$ 0.0328\\
\hline
\end{tabular}
\caption{Ablation study on contrastive loss}
\label{table: ablation contrastive loss}
\end{table}

For each site representation $\mathbf{z}_{s_i,t}$ from site $s_i$ at time $t$, we define the Positive instances and Negative instances as the same-site neural representation and the different sites neural representation. For simplicity, we first average the representation for all time $t$. The temperature $\tau$ controls the smoothness of the similarity distributions; we set $\tau = 0.1$ consistently across all experiments on the NPCS dataset. Thus, we contrast representations from different sites:
\begin{equation}
\mathcal{L}_{\text{contrast}} = -\frac{1}{S} \sum_{i=1}^{S} \log \frac{\exp(\text{sim}(\bar{\mathbf{z}}_i, \bar{\mathbf{z}}_i)/\tau)}{\sum_{j=1}^{S} \exp(\text{sim}(\bar{\mathbf{z}}_i, \bar{\mathbf{z}}_j)/\tau)}
\end{equation}
We showed the ablation study for the contrastive loss in Table~\ref {table: ablation contrastive loss}. Our results showed that contrastive loss further encourages the RPNT model to learn more robust neural representations, resulting in better downstream decoding performance. 

\section{Compare with Baseline Models}
\label{Supplementary: Compare with Baseline Models}
We compared our RPNT with current decoding models as baselines. Specifically, we included the Wiener filter~\cite{van2018signal} and several standard machine learning models such as MLP~\cite{glaser2020machine}, GRU~\cite{cho2014learning}, S4D~\cite{gu2021efficiently}, Mamba~\cite{gu2023mamba}, Transformer~\cite{vaswani2017attention,devlin2019bert} that were trained from scratch on the downsrteam single session. We also included two baseline pretrained models: NDT2~\cite{ye2023neural} and the latest POSSM~\cite{ryoo2025generalizable}. We acknowledge the author's contribution~\cite{ryoo2025generalizable} for implementing and testing all the baseline models in the main paper Table 1.

\section{Sweeping analysis for RPNT model}
\label{Supplementary: Sweeping analysis for RPNT model}
\begin{table}
\centering
\begin{tabular}{c|c}
\hline
Transformer layers & Cross-Task (T-RT) \\
\hline 
2 & 0.8482 $\pm$ 0.0955  \\
3 & 0.8432 $\pm$ 0.1004  \\
4 & 0.8515 $\pm$ 0.1071  \\
5 & 0.8431 $\pm$ 0.0914  \\
6 & 0.8461 $\pm$ 0.0942  \\
\hline
\end{tabular}
\caption{Sweeping study on transformer layer}
\label{table: Sweeping study on transformer layer}
\vspace{0.5cm}
\centering
\begin{tabular}{c|c}
\hline
Attention heads & Cross-Task (T-RT) \\
\hline 
4  & 0.8320 $\pm$ 0.1054  \\
8  & 0.8531 $\pm$ 0.0931  \\
16 & 0.8515 $\pm$ 0.1071  \\
32 & 0.8509 $\pm$ 0.1027  \\
64 & 0.8505 $\pm$ 0.0935  \\
\hline
\end{tabular}
\caption{Sweeping study on attention head}
\label{table: Sweeping study on attention head}
\vspace{0.5cm}
\centering
\begin{tabular}{c|c}
\hline
Kernel size & Cross-Task (T-RT) \\
\hline 
$[3, 3]$ & 0.8317 $\pm$   0.0945 \\
$[7, 7]$ & 0.8481 $\pm$   0.0954 \\
$[9, 9]$ & 0.8515 $\pm$   0.1071 \\
$[11, 11]$ & 0.8411 $\pm$ 0.0966 \\
$[15, 15]$ & 0.8479 $\pm$ 0.0900 \\
\hline
\end{tabular}
\caption{Sweeping study on kernel size}
\label{table: Sweeping study on kernel size}
\end{table}
We showed the sweeping analysis result for our RPNT model on the T-RT task, including temporal transformer layers (see table~\ref{table: Sweeping study on transformer layer}), attention heads (see table~\ref{table: Sweeping study on attention head}). We further showed the sweeping analysis of the kernel size  (see table~\ref{table: Sweeping study on kernel size}), which indicates the data-driven optimal context historical window was around 180ms.

\section{Comparison of two causal masking implementations}
\label{Supplementary: Comparison of two causal masking implementations}
\begin{table}
\centering
\begin{tabular}{l|c|c}
\hline
Method & Cross-Task (T-RT) & Cross-Site (B-CS) \\
\hline 
Causal masking before the convolution& 0.8474 $\pm$ 0.0902 & 0.6617 $\pm$ 0.019 \\
Causal masking after the convolution &  0.8515 $\pm$ 0.1071 & 0.6612 $\pm$ 0.0328\\
\hline
\end{tabular}
\caption{Comparison of two causal masking implementations.}
\label{table: Comparison of two causal masking implementations}
\end{table}
We showed the results on the T-RT and B-CS tasks (as they are harder compared to the C-CO and T-CO tasks) using two causal masking implementations (see table~\ref{table: Comparison of two causal masking implementations}). Our results did not show obvious performance differences in both implementations. 

\section{Spatial attention map visualization in NPCS dataset}
\label{Supplementary: Visualization}
\begin{figure}
    \centering
    \includegraphics[width=1.0\linewidth]{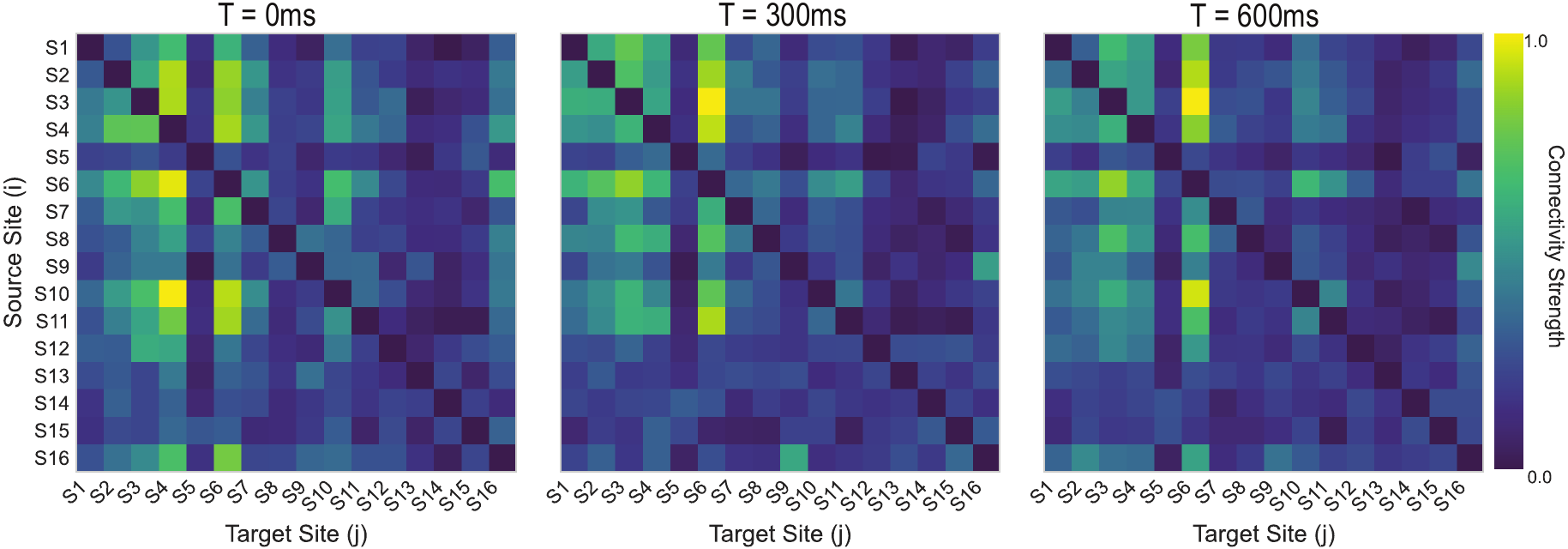}
    \caption{Spatial attention map provided data-driven brain network insights for motor behaviors.}
    \label{Appendix figure 1: Attention map visualization}
\end{figure}
We examined the attention-based FC map $\mathbf{C} \in \mathbb{R}^{T \times S \times S}$ from the spatial encoder for the Neuropixel dataset. Appendix figure~\ref{Appendix figure 1: Attention map visualization} showed the time evolution (0ms, 300ms, 600ms) of FC during the reaching period. We observed that site 6 remained quite active, whereas large site numbers, e.g., S10 and later, seemed inactive. Further, it revealed a small network interaction within the large PMd-M1 region, which provided data-driven insights into neural mechanisms of motor behavior. \textcolor{black}{These results might provide insight for future intervention investigation of brain sites to verify the data-driven observations.}

\section{Hyperparameters and Computational resources}
\label{Appendix: hyperparameters}
We showed the detailed hyperparameter setup for the RPNT model pretraining and SFT in table~\ref{table: Hyperparameter setup for pretraining RPNT} and table~\ref{table: Hyperparameter setup for SFT of RPNT}, respectively. Pretraining of RPNT was trained using a single A40 GPU, consuming 3GB and 9GB of GPU memory for the public dataset and neuropixel dataset, respectively, which takes around 12 hours for training. We saved the best checkpoint based on the early stopping criteria for the downstream SFT. For the LRTCH dataset, it ran 2 hours for C-CO, T-CO, and T-RT tasks. For the NPCS dataset, it ran 15 minutes for B-CS tasks. The SFT results were reported based on the last epoch.

\begin{table}
\centering
\begin{tabular}{l|l|l}
\hline 
\textbf{Parameter} & \textbf{Public dataset (LTRCH)} & \textbf{Neuropixel dataset (NPCS)} \\
\hline
Model dimension & 512 & 384 \\
Temporal layers & 4 & 4\\
Spatial layers & N/A & 2\\
Attention heads & 16 & 12\\
Kernel size & $[9,9]$ & $[9,9]$\\
Dropout rate & 0.1 & 0.1\\
Batch size & 64 & 32\\
Epoch & 100 & 100 \\
Warm-up epoch & 50 & 10 \\
Weight decay & 0.01 & 0.01 \\
Learning rate & $5 \times 10^{-5}$ & $5 \times 10^{-5}$\\
Gradient clip & 1.0 & 1.0\\
Contrastive loss ($\lambda$) & N/A & 0.1\\
Random seed & 3407 & 3407 \\
\hline
\end{tabular}
\caption{Hyperparameter setup for pretraining RPNT on both dataset}
\label{table: Hyperparameter setup for pretraining RPNT}
\end{table}

\begin{table}[h]
\centering
\begin{tabular}{l|l|l}
\hline 
\textbf{Parameter} & \textbf{Public dataset (LTRCH)} & \textbf{Neuropixel dataset (NPCS)} \\
\hline
Dropout rate & 0.1 & 0.1\\
Batch size & 32 & 32\\
Epoch & 200 & 200 \\
Weight decay & 0.01 & 0.01 \\
Learning rate & $1 \times 10^{-4}$ & $1 \times 10^{-4}$\\
Gradient clip & 1.0 & 1.0\\
Random seed & 3407 & 3407 \\
\hline
\end{tabular}
\caption{Hyperparameter setup for SFT of RPNT on both datasets}
\label{table: Hyperparameter setup for SFT of RPNT}
\end{table}

\end{document}